%% file: template.tex
\documentclass{article}

\usepackage[eandd,final]{neurips_2026}

\usepackage[utf8]{inputenc} 
\usepackage[T1]{fontenc}    
\usepackage{amsmath}
\usepackage{amsthm}
\usepackage{url}            
\usepackage{booktabs}       
\usepackage{amsfonts}       
\usepackage{nicefrac}       
\usepackage{microtype}      
\usepackage[table]{xcolor}  
\usepackage{mdframed}
\usepackage{graphicx}
\usepackage{float}
\usepackage{placeins}
\usepackage{array}
\usepackage{tabularx}
\usepackage{multirow}
\usepackage{hyperref}       
\usepackage{doi}

\definecolor{defbg}{RGB}{245,248,255}
\definecolor{defborder}{RGB}{110,140,255}
\newmdenv[
  backgroundcolor=defbg,
  linecolor=defborder,
  linewidth=0.6pt,
  roundcorner=2pt,
  innertopmargin=6pt,
  innerbottommargin=6pt,
  innerleftmargin=8pt,
  innerrightmargin=8pt,
  skipabove=6pt,
  skipbelow=6pt
]{definitionbox}

\theoremstyle{definition}
\newtheorem{definition}{Definition}

\newcommand{\msys}{\mathcal{M}}

\definecolor{revisionblue}{RGB}{0,72,160}

\title{When Stored Evidence Stops Being Usable: Scale-Conditioned Evaluation of Agent Memory}

\author{%
  Jiaqi Shao\textsuperscript{1,2}\thanks{Both authors contributed equally to this work and share first authorship. The author order is alphabetical.}\quad
  Yiyi Lu\textsuperscript{1}\footnotemark[1]\quad
  Yunzhen Zhang\textsuperscript{1}\quad
  Bing Luo\textsuperscript{1}\thanks{Corresponding author: \href{mailto:bl291@duke.edu}{\texttt{bl291@duke.edu}}.} \\
  \textsuperscript{1}Duke Kunshan University, China \\
  \textsuperscript{2}The Hong Kong University of Science and Technology, Hong Kong
}

\hypersetup{
hidelinks,
pdftitle={When Stored Evidence Stops Being Usable: Scale-Conditioned Evaluation of Agent Memory},
pdfauthor={Jiaqi Shao, Yiyi Lu, Yunzhen Zhang, Bing Luo},
pdfkeywords={agent systems, memory evaluation, scalability, evaluation protocols},
}

\begin{document}
\maketitle

\begin{abstract}
Memory-agent evaluations report fixed-snapshot accuracy or retrieval quality, but these scores do not show whether evidence remains usable as irrelevant sessions (sessions not annotated as task-relevant evidence for the query) accumulate.
We present a scale-conditioned evaluation protocol for agent memory under evidence-preserving growth: for each query, task evidence is held fixed while irrelevant sessions are added.
The protocol logs agent--memory trajectories and reports four diagnostics: budget-compliant reliability, tail memory-call burden, failure-regime decomposition, and the usable-scale boundary where reliability falls below the target.
Applied to LongMemEval and LoCoMo across flat, planar, and hierarchical memory interfaces, the protocol shows reliability loss is not a single phenomenon.
On LongMemEval, HippoRAG stays within the two-call budget but loses 16--20 percentage points in budget-compliant reliability as irrelevant sessions are added; LiCoMemory's observed failures depend strongly on the agent, with Qwen3-8B exceeding the budget while Qwen3-32B and Qwen3-235B remain reliable in the tested range.
The result supports a framework for making scalable-memory claims conditional on agent, interface, scale range, and interaction budget.
Code and evaluation artifacts: \url{https://github.com/AndyLu666/memory-at-scale}.
\end{abstract}

\section{Introduction}

LLM agents combine model reasoning with external tools, structured workflows, and autonomous environment interaction~\citep{toolformer,react,autogen,webarena}.
To support long-horizon interaction, agents are often equipped with memory systems that persist information across sessions and make relevant context available at inference time~\citep{generative_agents,memgpt,memorybank,amem,memos}.

Prior work on agent memory primarily falls into two categories. \emph{Systems} research designs memory backends that retrieve relevant evidence~\citep{lightmem,memos,hipporag,licomemory}. \emph{Evaluation} research typically assesses accuracy or retrieval quality using a fixed memory snapshot~\citep{longmemeval,locomo,memoryarena}. However, deployed agents accumulate sessions and observations over time~\citep{memgpt,longmemeval,locomo,hu2026memoryageaiagents}. A fixed-snapshot score cannot tell us how far a scalability claim extends as irrelevant accessible history grows, nor whether a later failure comes from retrieval ranking, excessive interaction, answer synthesis, or scoring noise.

\begin{figure}[t]
  \centering
  \includegraphics[width=0.9\linewidth]{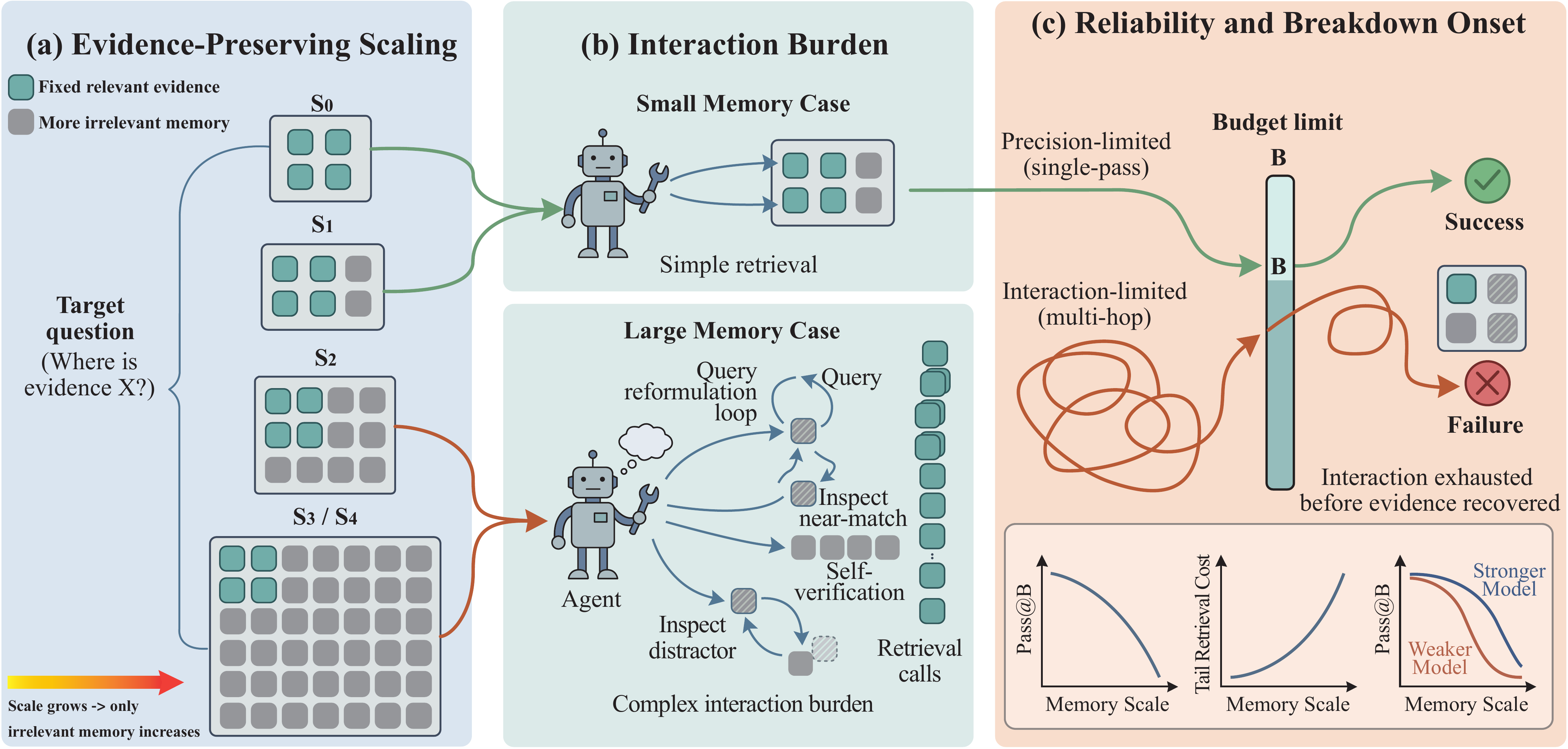}
  \caption{\textbf{Overview of memory at scale.}
(a) \textit{Evidence-preserving scaling}: task-relevant evidence is held fixed while irrelevant accessible memory grows.
(b) \textit{Interaction burden}: larger memories can trigger long-tail retrieve--verify loops (more agent-issued memory calls).
(c) \textit{Reliability and usable-scale boundary}: under a fixed retrieval budget, interaction burden and within-budget errors translate into failures, shifting the scale where reliability first falls below the target across agent--memory systems.}
  \label{fig:overview}
  \end{figure}

To address this gap, we study agent behavior under \textbf{memory at scale}: increasing irrelevant accessible memory while keeping task-relevant evidence fixed.
We adopt this evidence-preserving scaling construction from LongMemEval~\citep{longmemeval}.
Our goal is not to introduce a new benchmark or memory architecture, but to turn this construction into a trajectory-level evaluation protocol for agent--memory systems.
As illustrated in Figure~\ref{fig:overview}(a), this gives a controlled intervention for asking whether stored evidence remains usable as accessible history grows.

As accessible memory grows, the correct evidence can become harder to locate, select, verify, and use through a particular interface.
We use \emph{agent-facing interaction burden} to refer to this inference-time burden, distinct from storage size, offline indexing cost, hidden backend traversal, or end-to-end systems cost.
The form of the difficulty can depend on the memory interface: single-pass systems may answer incorrectly within budget, whereas multi-hop systems may spend too many memory calls on traversal and verification (Figure~\ref{fig:overview}(b)).

We study memory scalability as an evaluation claim under evidence-preserving
growth: the memory-side intervention enlarges accessible history, but the
reported quantities are computed from agent--memory trajectories.
The protocol logs these trajectories and reports $\mathrm{Pass@B}$, tail
retrieval-call burden, failure-regime decomposition into $p_{\mathrm{exh}}$
and $p_{\mathrm{wrong}}$, and a usable-scale boundary marking where reliability
first falls below the target.
Our central claim is that scalable-memory reports should be stated as agent--interface claims tied to explicit scale and retrieval-budget conditions, not only as storage capacity or fixed-snapshot accuracy.

\textbf{Scope of the evaluation.}
We evaluate usable memory under evidence-preserving growth.
The retrieval-call budget counts agent-issued memory API calls visible in the trajectory.
We do not claim to measure total storage scalability, offline indexing cost, hidden backend graph computation, semantic density of returned evidence, adversarial memory robustness, or stale/contradictory memory management.
Returned evidence units are matched in count, but the comparison should be read as an agent-facing interface evaluation rather than a full end-to-end systems-cost comparison.
Our contributions are summarized:
\begin{itemize}
    \item \textbf{A scale-conditioned evaluation protocol for usable agent memory.}
    For each query, annotated task evidence is fixed while irrelevant sessions are injected according to a shared scale ladder. The protocol evaluates whether an agent can still use the evidence through a specified memory interface under a stated memory-call budget.

    \item \textbf{A trajectory-level diagnostic suite for scalable-memory claims.}
    We report budget-compliant reliability, tail retrieval-call burden, failure-regime decomposition, and scale boundary where reliability falls below the target. Together, these diagnostics distinguish systems that answer incorrectly within budget from systems that violate interaction budget.

    \item \textbf{An empirical audit across memory-interface families and agents.}
    On LongMemEval and LoCoMo, we apply protocol to flat, planar, and hierarchical memory interfaces. The results show that similar reliability curves can correspond to different observed failure regimes, and scalable-memory claims must be tied to the agent, interface, scale range, and budget.
\end{itemize}

\section{Related Work}

\paragraph{Memory scale and memory interfaces.}
Prior work studies memory scale as long input context, growing conversational history, and external memory-system design. Long-context benchmarks show that relevant evidence can be underused even when it appears in the input~\citep{lost_middle,longbench,ruler}; long-memory and agent-memory benchmarks evaluate growing histories, multi-session conversations, and interactive memory use~\citep{longmemeval,locomo,hu2025memoryagentbench,memoryarena,anatomy_agentic_memory}. Memory systems organize external state through OS-like, lifecycle-managed, lightweight, graph, or hierarchical interfaces~\citep{memgpt,memos,lightmem,hipporag,licomemory}. Following recent taxonomies, we use flat, planar, and hierarchical interfaces as the retrieval-time interface axis~\citep{hu2026memoryageaiagents}. These works establish that scale and interface design matter, but they do not by themselves make clear whether scale-induced failures come from evidence selection, excessive agent-memory interaction, stopping behavior, or answer synthesis.

\paragraph{Auditing memory-scalability claims.}
Endpoint accuracy, retrieval recall, latency, and cost are useful, but they can collapse distinct failure regimes into one score. Under evidence-preserving growth, a reliability drop may mean that evidence was not returned, that the agent used too many memory calls, that search failed to stop, or that the final answer was wrong despite staying within budget. We therefore treat memory scalability as an agent--interface evaluation claim: with task evidence fixed, irrelevant memory increased, retrieval-call budget stated, and returned evidence exposure matched, a scalable-memory report should include $\mathrm{Pass@}B$, $\mathrm{P90R}$, budget-induced versus wrong-within-budget rates, and breakdown onset. Our goal is not another memory-system leaderboard, but an audit lens for asking whether stored evidence remains usable as accessible memory grows.

\section{Scale-Conditioned Agent--Memory Evaluation}
\label{sec:method}

This section specifies the evaluation protocol and trajectory-level diagnostics used to make scalable-memory claims auditable.

\begin{definitionbox}
\noindent\textbf{Protocol.}
For each benchmark query $q$, the evaluation proceeds as follows:
\begin{enumerate}
    \item identify the annotated task-relevant evidence sessions $E(q)$;
    \item construct histories $\mathcal{H}^{(s)}(q)$ by holding $E(q)$ fixed and adding $N_{\mathrm{irr}}(s)$ irrelevant sessions;
    \item run agent $A$ with memory interface $\msys$ and log all agent-issued memory calls;
    \item score the final answer and compute budget-compliant reliability, retrieval-call quantiles, failure-regime decomposition, and breakdown onset.
\end{enumerate}
\end{definitionbox}

\subsection{Agent--Memory Interaction Under Scale}
\label{sec:method:interaction}

We study usable memory scalability as an agent--interface operating property,
not as intrinsic storage scalability of the memory backend alone. The scale
intervention is applied to the accessible history $\mathcal H^{(s)}(q)$, while
the measured object is the trajectory induced when an agent model $A$ uses a
memory interface $\mathcal M$ under a retrieval-call budget. For a query
$q \sim \mathcal D$ and scale condition $s \in \mathcal S$, let
$\mathcal H^{(s)}(q)$ denote the accessible history at scale $s$. We write
$\mathcal H^{(s)}$ when the context is clear.

\paragraph{Evidence-preserving memory scaling.}
Let $y(q)$ denote the gold answer and let $E(q)$ denote the set of task-relevant evidence sessions for $q$. A family $\{\mathcal{H}^{(s)}(q)\}_{s\in\mathcal{S}}$ is \emph{evidence-preserving} if for every $s$ we have $E(q)\subseteq \mathcal{H}^{(s)}(q)$, the set $E(q)$ is unchanged across $s$, and $\mathcal{H}^{(s)}(q)\setminus E(q)$ contains only additional irrelevant sessions.
This follows LongMemEval~\citep{longmemeval}. It keeps answer evidence fixed while changing the retrieval-control problem.

\paragraph{Agent--memory trajectories.}
Let $\mathcal{A}_{\mathrm{ext}}(\msys)$ denote the set of agent-issued external memory API calls that count toward the retrieval-call budget. A rollout is
\[
\tau = (o_1, a_1, o_2, a_2, \dots, o_T, a_T), \quad \tau \sim P\!\left(\tau \mid q, A, \msys, \mathcal{H}^{(s)}(q)\right),
\]
where $o_t$ and $a_t$ denote observations and actions, and $t$ indexes steps. Define the retrieval-call count:
\[
R(\tau) := \sum_{t=1}^{T}\mathbf{1}\!\left[a_t \in \mathcal{A}_{\mathrm{ext}}(\msys)\right],
\]
and let $C(\tau)\in\{0,1\}$ denote final-answer correctness: $C(\tau)=1$ if rollout $\tau$ returns a correct final answer, and $C(\tau)=0$ otherwise.
In our experiments, $\mathcal{S}=\{s_0,\dots,s_4\}$ is a shared discrete ladder with matched irrelevant-session counts across benchmarks (Table~\ref{tab:s_longmemeval}). Budget IDs and the returned-item parity constraint are specified in \S~\ref{sec:exp}. For graph-based systems, we use top-$k=12$, and a returned item is a model-visible chunk-level evidence unit, not an internal graph node, path, or provenance session ID.

\subsection{Trajectory-Level Diagnostics for Memory Scalability}
\label{sec:metrics}

To characterize agent--memory behavior as $s$ increases, we ask: (1) Does reliability hold under a fixed retrieval-call budget? (2) Does interaction burden grow with scale? (3) What observed failure source accounts for any reliability drop? (4) At what scale does reliability break down?

\subsubsection{Does reliability hold under a fixed retrieval-call budget?}

\begin{definitionbox}
\begin{definition}[Budget-compliant reliability --- $\mathrm{Pass@B}$]
\label{def:passatb}
Let $R(\tau)$ be the retrieval-call count defined above. Define the per-rollout success indicator
\[
\mathrm{pass}_B(\tau) := \mathbf{1}\!\left[C(\tau)=1 \;\wedge\; R(\tau)\le B\right].
\]
Budgeted reliability under system $(A,\msys)$ at scale $s$ is
\begin{equation}
\mathrm{Pass@B}(A,\msys,s)
:= \mathbb{E}_{q\sim\mathcal{D}}\,
\mathbb{E}_{\tau\sim P(\cdot\mid q,A,\msys,\mathcal{H}^{(s)}(q))}
\!\left[\mathrm{pass}_B(\tau)\right].
\label{eq:passatb}
\end{equation}
\end{definition}
\end{definitionbox}

$\mathrm{pass}_B(\tau)\in\{0,1\}$ is a per-rollout binary indicator, and $\mathrm{Pass@B}$ is its expectation over the joint distribution of tasks and rollouts. In practice, we estimate $\mathrm{Pass@B}$ by averaging $\mathrm{pass}_B(\tau)$ over evaluated rollouts.
This metric is a budget-compliant reliability measure, not pass@$k$ from code-generation evaluation: a rollout counts only if the final answer is correct and retrieval-call count is at most $B$.
Under a stated budget and scale range, a scalability claim should specify if $\mathrm{Pass@B}$ is maintained from $s_0$ to larger $s$ values, rather than reporting only the evidence-only condition.

\subsubsection{Does interaction burden grow with scale?}
$\mathrm{Pass@B}$ captures the outcome but not its cause: a drop could arise from heavier retrieval demand or from more within-budget errors, and both compress $\mathrm{Pass@B}$ equally. We therefore track how the retrieval-call distribution shifts with $s$.

\begin{definitionbox}
\begin{definition}[Retrieval-call quantile --- $Q_p$]
\label{def:burden}
Let $P_{A,\msys,s}$ denote the marginal distribution of $R(\tau)$ under $q\sim\mathcal{D}$ and $\tau\sim P(\cdot\mid q,A,\msys,\mathcal{H}^{(s)}(q))$. Define the $p$-th quantile
\[
Q_p(A,\msys,s) := \inf\!\left\{r\in\mathbb{N}_0 : P_{A,\msys,s}\!\left(R(\tau)\le r\right)\ge p\right\}.
\]
\end{definition}
\end{definitionbox}

We use the quantile at $p=0.9$ to track tail interaction burden:
\begin{equation}
\mathrm{P90R}(A,\msys,s) := Q_{0.9}(A,\msys,s).
\label{eq:burden}
\end{equation}

$\mathrm{P90R}$ tracks the tail of the retrieval-call distribution, capturing long retrieve--verify episodes that arise under many irrelevant sessions. We focus on the tail because burden accumulates in hard queries first; the median remains low even when scale-induced failures are already frequent. (The median $\mathrm{MedR}:=Q_{0.5}$ is reported in Appendix~\ref{app:diagnostics} as a supplementary diagnostic.)

\subsubsection{What observed failure regime accounts for any reliability drop?}
Growing interaction burden does not identify the failure regime. A trajectory with $R(\tau)>B$ is budget-induced with the stated online interaction budget; it is not necessarily intrinsically unanswerable. Conversely, a trajectory may stay within budget and still return a wrong final answer. We therefore decompose observed failures into two reported masses: wrong-within-budget answers and budget-induced trajectories.

\begin{definitionbox}
\begin{definition}[Failure-regime decomposition]
\label{def:decomp}
Define
\begin{equation}
\begin{aligned}
p_{\mathrm{wrong}} &:= \Pr_{q,\tau}\!\left[C(\tau)=0 \wedge R(\tau)\le B\right],\\
p_{\mathrm{exh}} &:= \Pr_{q,\tau}\!\left[R(\tau)>B\right].
\end{aligned}
\label{eq:decomp_terms}
\end{equation}
with $(q,\tau)$ distributed according to $q\sim\mathcal{D}$ and $\tau\sim P(\cdot\mid q,A,\msys,\mathcal{H}^{(s)}(q))$.
\end{definition}
\end{definitionbox}

The same $\mathrm{Pass@B}$ drop can arise from different observed regimes, making this decomposition essential for interpreting what an aggregate score supports.
We report $p_{\mathrm{exh}}$ and $p_{\mathrm{wrong}}$ alongside $\mathrm{Pass@B}$.
These are operational diagnostic categories: a wrong-within-budget answer may reflect retrieval ranking, returned-evidence formatting, answer synthesis, ambiguity, or judging error.

\subsubsection{At what scale does reliability break down?}
To summarize how far a memory-scalability claim extends, we report the first memory scale where reliability falls below a threshold.

\begin{definitionbox}
\begin{definition}[Breakdown onset --- $s^*_\alpha(A,\msys;B)$]
\label{def:onset}
Fix a reliability threshold $\alpha\in(0,1)$. If reliability falls below $\alpha$ at any evaluated scale, let $s_j$ be the first such scale in $\{s_0,\dots,s_4\}$. We report breakdown onset as the number of added irrelevant sessions at that scale:
\begin{equation}
s^{*}_{\alpha}(A,\msys;B) := N_{\mathrm{irr}}(s_j).
\label{eq:onset}
\end{equation}
If no such scale exists, we write $s^{*}_{\alpha}(A,\msys;B)>N_{\mathrm{irr}}(s_4)$.
\end{definition}
\end{definitionbox}

A smaller $s^*_\alpha(A,\msys;B)$ indicates earlier loss of usable scale under the same budget. We report $s^*$ in units of added irrelevant sessions, so $s^*=100$ means reliability first falls below threshold at $s_1$, and $s^*>400$ means no breakdown is observed through $s_4$. $s^*$ should always be read jointly with the failure decomposition: two systems with the same $s^*$ may reach the boundary through entirely different mechanisms. We set $\alpha=0.7$ in all primary results.
This threshold is an operating threshold, not a universal acceptability criterion (Appendix~\ref{app:diagnostics} reports multi-budget sensitivity).


\section{Experimental Setup}
\label{sec:exp}
\paragraph{Models and Memory Interfaces.}
We evaluate Qwen3-8B, Qwen3-32B, and Qwen3-235B\footnote{\texttt{Qwen3-235B-A22B-Instruct-2507}.}~\citep{qwen3} with flat (OpenClaw, MemOS-text), planar (Mem0, MemOS-Tree), and hierarchical (HippoRAG, LiCoMemory) memory interfaces. Model details, local serving-memory estimates, additional LLaMA-3.1 diagnostics, and the full interface taxonomy are in Appendix Tables~\ref{tab:model-tiers},~\ref{tab:local_compute_estimates},~\ref{tab:memory-taxonomy-full} and Appendix~\ref{app:llama}.
All systems use the same accessible history $\mathcal{H}^{(s)}$, retrieval-call budget, and number of model-visible returned evidence units at each scale; parity details are in Appendix~\ref{app:fp_parity}. The budget counts agent-visible memory API calls, not hidden backend traversal or indexing, and is applied post-hoc: rollouts are not truncated, but success requires $R(\tau)\le B$ (Eq.~\ref{eq:passatb}). Primary results use $B_0=2$; budget sensitivity and token-cost diagnostics are in Appendix~\ref{app:diagnostics}.
\paragraph{Benchmarks and Scale Ladder.}
\label{sec:budget_scale}
LongMemEval and LoCoMo share the scale ladder $\{s_0,\dots,s_4\}$ (Table~\ref{tab:s_longmemeval}): $s_0$ is evidence-only, and larger scales add $N_{\mathrm{irr}}(s)$ irrelevant sessions while keeping task evidence fixed. Thus changes across $s$ reflect agent--memory behavior under a larger accessible history, not removal of evidence. LoCoMo external validity, evaluation sizes, and confidence intervals are reported in Appendix~\ref{app:locomo_external} (Table~\ref{tab:locomo_r2}) and Appendix~\ref{app:evaluation_size}.

For each trajectory, we use an LLM-as-judge scorer for the final answer, aligned with the MemOS evaluation protocol~\citep{memos}. The judge prompt, model setting, and conversion to $C(\tau)$ are detailed in Appendix~\ref{app:llm_judge}.

\FloatBarrier


\bibliographystyle{unsrtnat}

\section{Results}
\label{sec:experiments}
\paragraph{Overview.}
We organize the results around what the protocol diagnoses. First, memory growth changes budgeted reliability even under a controlled scale ladder that keeps the task evidence fixed (\S~\ref{sec:main_longmemeval}). Second, failure decomposition separates distinct observed failure sources: within-budget wrong answers, search-and-stopping instability, and budget exhaustion under weak interface control (\S~\ref{sec:f2_f3_mechanism}). Third, model scale is associated with different memory-control behavior on the multi-hop interface, reducing tail retrieval calls and budget exhaustion in the evaluated runs (\S~\ref{sec:results_control}). Finally, breakdown onset summarizes these diagnostics as a usable-scale boundary (\S~\ref{sec:results_breakdown}).

\begin{figure*}[t]
  \centering
  \includegraphics[width=0.88\textwidth]{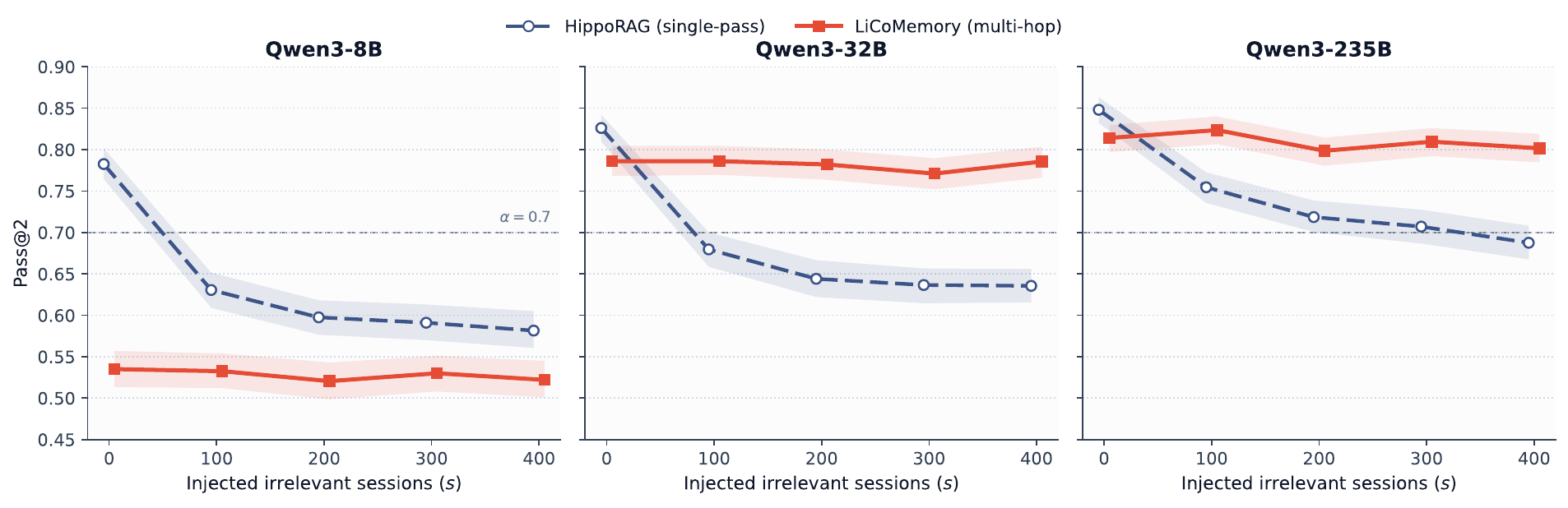}
\caption{\textbf{Usable memory scalability is a joint operating property of the agent, the memory interface, the scale range, and the interaction budget.}
HippoRAG loses reliability as memory scales; LiCoMemory exhibits two distinct patterns: Qwen3-8B is already below the reliability threshold at $s_0$, whereas Qwen3-32B/Qwen3-235B remain reliable through the largest tested scale with breakdown onset $s^{*}_{0.7}(A,\msys;B_0)>400$ added irrelevant sessions (Eq.~\ref{eq:onset}).
$\mathrm{Pass@}B_0$ (Eq.~\ref{eq:passatb}) is reported as mean $\pm$ 95\% CI; the dashed line marks $\alpha=0.7$.}
\label{fig:main_longmemeval_pass}
\end{figure*}

\subsection{Usable memory scalability is a joint operating property of the agent, the memory interface, the scale range, and the interaction budget}
\label{sec:main_longmemeval}

\begin{table}[t]
\centering
\small
\caption{\textbf{Reliability at scale depends jointly on memory interface and agent capability (LongMemEval, $B_0=2$).}
Rows report $\mathrm{Pass@}B_0$ at $s_0/s_4$, $\mathrm{P90R}$ at $s_4$, and breakdown onset---added irrelevant sessions before reliability falls below 70\% (\textbf{>400}: no observed breakdown). Mean $\pm$ 95\% CI. Onset uses point estimates, with CIs shown separately for task-level uncertainty.}
\tiny
\label{tab:main_r1}
\resizebox{\textwidth}{!}{%
\input{generated/longmemeval/table_r1_longmemeval.tex}
}
\end{table}

Figure~\ref{fig:main_longmemeval_pass} and Table~\ref{tab:main_r1} establish the first high-level finding: memory growth changes the agent-facing problem, not just the amount of stored data. The scale ladder holds the required evidence fixed and only adds irrelevant sessions, so changes in $\mathrm{Pass@}B_0$ measure the burden of retrieving, selecting, verifying, and using evidence through a particular memory interface.
This controlled contrast argues against a purely storage-centric interpretation of the reliability curve. Reliability can change even when annotated evidence remains present, because the agent must operate over a larger accessible history while using the same interface and retrieval-call budget.
Appendix Figure~\ref{fig:app_qwen_claim_i} gives the corresponding endpoint and retrieval-tail summaries.

{\bf Scalability is a joint operating property of the agent and memory interface.} HippoRAG starts from a reliable $s_0$ operating point, yet drops by 20.0, 19.1, and 16.0 $\mathrm{Pass@}B_0$ points from $s_0$ to $s_4$, pushing even the strongest model below the 70\% threshold. Conversely, LiCoMemory with Qwen3-8B changes only from 53.5\% to 52.2\%, but is already below threshold at $s_0$. In contrast, Qwen3-32B and Qwen3-235B with LiCoMemory remain above threshold across all tested scales ($78.6\% \rightarrow 78.5\%$ and $81.4\% \rightarrow 80.1\%$). These outcomes illustrate why a memory-scalability claim should specify the agent, interface, scale range, and call budget.
The next question is not only whether reliability changes, but how it changes.

\subsection{Finding 2: Similar reliability drops hide different failure regimes}
\label{sec:results_burden}
\label{sec:f2_f3_mechanism}
We next ask what burden produces the reliability curves in Figure~\ref{fig:main_longmemeval_pass}. We inspect the retrieval-call tail through $\mathrm{P90R}$ (Table~\ref{tab:main_r1}) and the observed failure source through $p_{\mathrm{exh}}$ and $p_{\mathrm{wrong}}$ (Table~\ref{tab:main_r1b}).
These categories are operational diagnostics. Budget exhaustion means the rollout would violate the stated retrieval-call constraint, not that the question is intrinsically unanswerable. Wrong-within-budget failures may arise from retrieval ranking, returned-evidence formatting, answer synthesis, ambiguity, or judging error. The point is to localize where finite-budget agent--memory interaction breaks, rather than to assign every error to a single subsystem.

\begin{table}[t]
\centering
\small
\caption{\textbf{Failure-regime decomposition for evaluated memory systems under $B_0=2$.}
Columns are grouped by budget-induced trajectories ($p_{\mathrm{exh}}=\Pr[R>B]$) versus wrong-within-budget answers ($p_{\mathrm{wrong}}$); red boxes mark the largest observed failure category at $s_4$.}
\label{tab:main_r1b}
\tiny
\resizebox{\columnwidth}{!}{\input{generated/longmemeval/table_r1b_longmemeval_failure.tex}}
\end{table}

\begin{figure*}[t]
\centering
\includegraphics[width=0.78\textwidth]{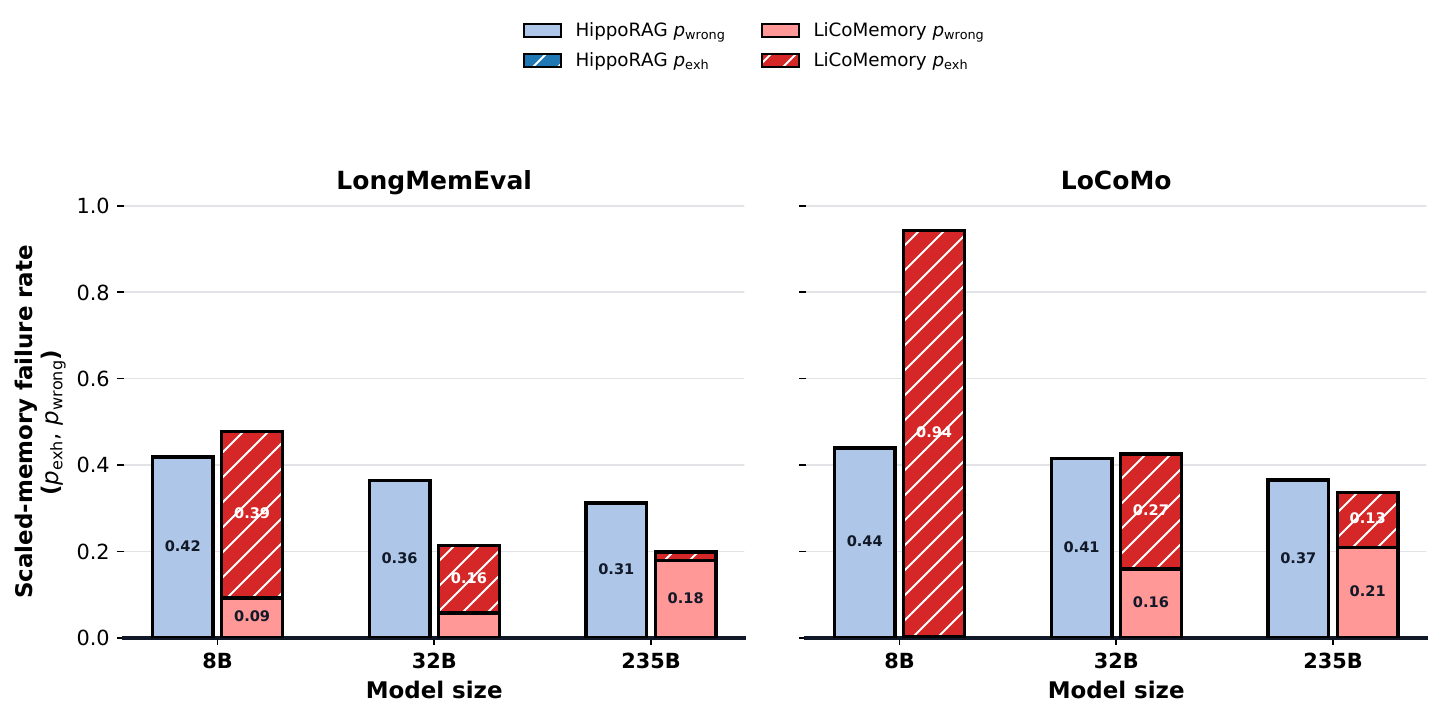}
\caption{\textbf{Failure-regime decomposition at $s_4$ ($B_0=2$).}
Failure probability is decomposed into budget-induced trajectories ($p_{\mathrm{exh}}$) and within-budget wrong answers ($p_{\mathrm{wrong}}$; Eq.~\ref{eq:decomp_terms}).}
\label{fig:main_failure_decomp}
\end{figure*}

The observed regime differs by interface: some systems fail within $B_0$, while others exhibit a control burden where retrieval and verification exceed the call constraint.

{\bf Flat interfaces keep access simple but shift burden to evidence use.} In OpenClaw, at $s_4$, $p_{\mathrm{exh}}$ is near zero, while $p_{\mathrm{wrong}}$ remains high across Qwen3-8B, Qwen3-32B, and Qwen3-235B (54.5\%, 42.0\%, and 27.5\%). Flat retrieval usually avoids long search loops, but at scale the returned context can be noisy, insufficient, or hard to synthesize, so the agent answers incorrectly within budget. More details are provided in Appendix Figures~\ref{fig:app_openclaw_scale_output} and~\ref{fig:app_openclaw_tail_cost}.

{\bf Among the evaluated planar adapters, Mem0 shows search-and-stopping instability.} Qwen3-8B mainly stops with a wrong answer at $s_4$ ($p_{\mathrm{wrong}}=85.6\%$), whereas Qwen3-32B and Qwen3-235B mostly produce budget-induced trajectories ($p_{\mathrm{exh}}=83.6\%$ and 79.2\%). Iterative planar search gives the agent more freedom, but that freedom creates a stopping burden: the agent may stop early with the wrong memory, or continue searching and verifying until it violates $B_0=2$.

{\bf Hierarchical interfaces split into two observed regimes.} HippoRAG behaves like single-pass structured retrieval: $p_{\mathrm{exh}}=0$ across Qwen tiers, but $p_{\mathrm{wrong}}$ rises from $s_0$ to $s_4$ (21.8\% to 41.8\%, 17.4\% to 36.5\%, and 15.2\% to 31.2\%). Operationally, this appears as a wrong-within-budget regime, consistent with stronger competition for the fixed set of returned evidence units; the decomposition does not require attributing every such answer to retriever precision alone. LiCoMemory instead exposes a multi-hop control burden: Qwen3-8B has many budget-induced trajectories at $s_4$ ($p_{\mathrm{exh}}=38.6\%$), but this burden falls with model scale.

Memory growth therefore does not induce one universal observed failure regime: flat retrieval mainly yields wrong-within-budget evidence-use failures, planar memory shows search-and-stopping instability, single-pass structured retrieval returns within-budget wrong answers, and multi-hop hierarchical retrieval can violate the budget when model-side control is weak. The result is not that hierarchy is uniformly better, but that scalable-memory claims need failure-regime reporting.

\subsection{Finding 3: Larger agents improve interface control only for some memory interfaces}
\label{sec:results_control}
In the evaluated LiCoMemory setting, larger Qwen3 agents show better budget compliance and lower tail retrieval-call burden. On LiCoMemory at $s_4$, the retrieval-call tail shrinks from $\mathrm{P90R}=5$ to 4 to 2 across Qwen3-8B, Qwen3-32B, and Qwen3-235B; over the same tiers, budget-induced trajectories fall from 38.6\% to 15.7\% to 2.0\%, while $\mathrm{Pass@}B_0$ rises from 52.2\% to 78.5\% to 80.1\% (Tables~\ref{tab:main_r1}--\ref{tab:main_r1b}).
This is more informative than a final-accuracy comparison alone: the memory interface, scale condition, and retrieval-call budget are fixed, while the tail-call statistic and budget-induced rate change with the agent model. The improvement therefore appears in observed search and stopping behavior, not only in the answer-level score.

\begin{figure*}[t]
\centering
\begin{minipage}{0.49\textwidth}
\centering
\includegraphics[width=\textwidth]{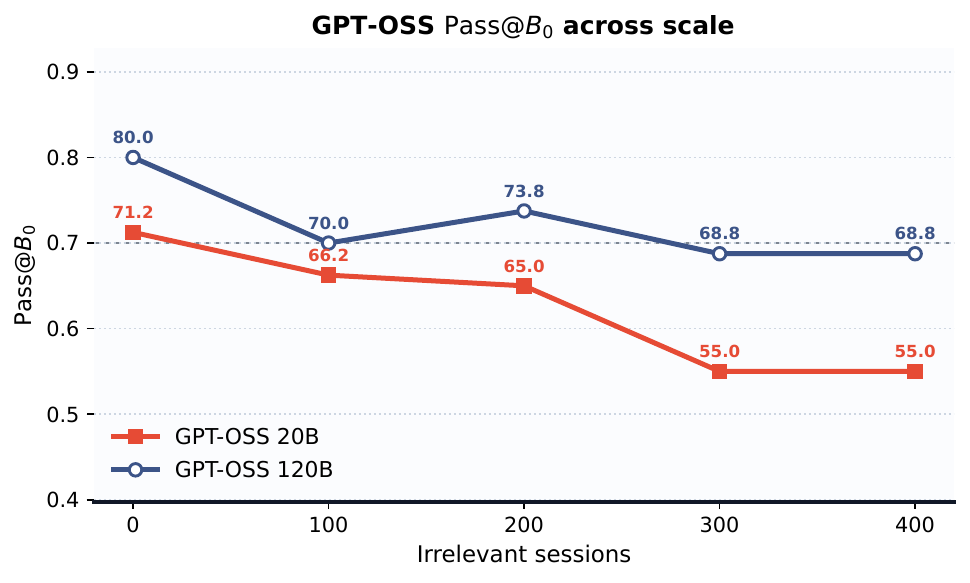}
\end{minipage}
\hfill
\begin{minipage}{0.49\textwidth}
\centering
\includegraphics[width=\textwidth]{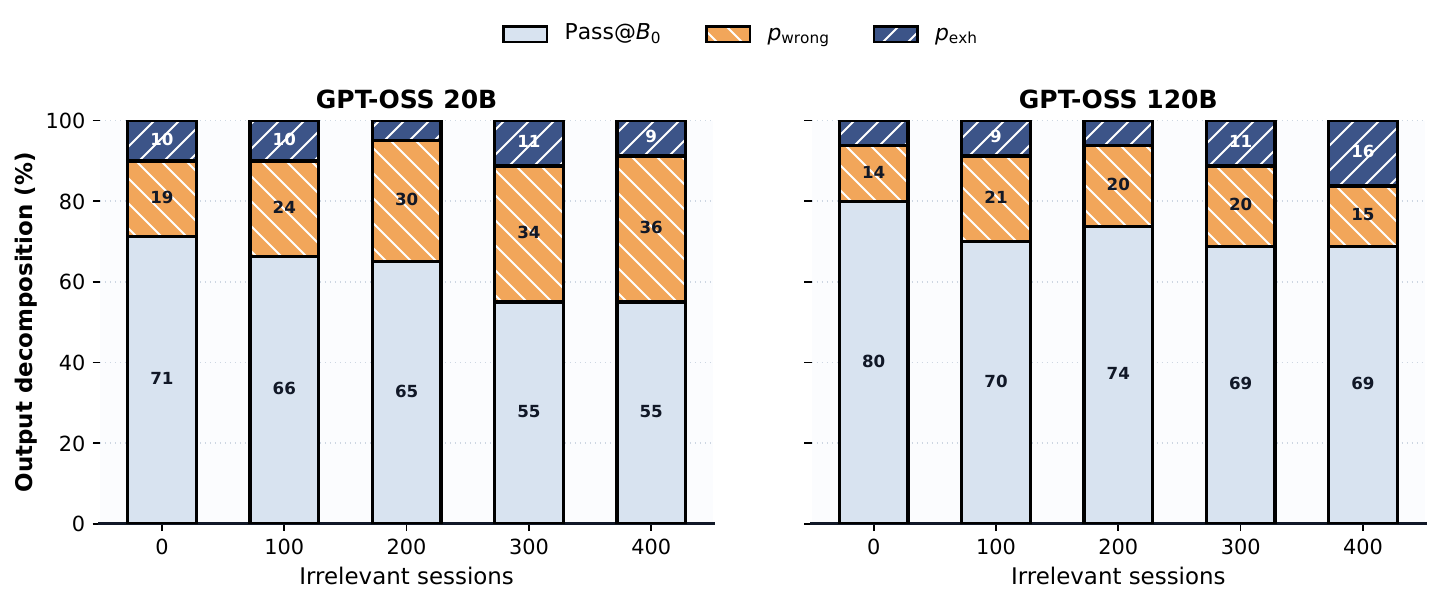}
\end{minipage}
\caption{\textbf{GPT-OSS LiCoMemory under the same scale-and-budget protocol.}
Left: $\mathrm{Pass@}B_0$ across the LongMemEval scale ladder under $B_0=2$.
Right: output decomposition across scale separates $\mathrm{Pass@}B_0$, wrong-within-budget failures, and budget-induced trajectories.}
\label{fig:gptoss_licomemory_main}
\end{figure*}

Thus, in this interface, larger models more often stop within the retrieval-call constraint. Figure~\ref{fig:main_failure_decomp} visualizes the corresponding shift through the shrinking budget-induced loss on LiCoMemory. A matched trajectory in Appendix Figure~\ref{fig:qualitative_pair} illustrates the behavior on one item: the weaker model repeatedly retrieves and reformulates until budget exhaustion, whereas the stronger model reaches sufficient evidence after one effective retrieval.
Additionally, Figure~\ref{fig:gptoss_licomemory_main} adds GPT-OSS LiCoMemory as a model-family diagnostic under the same scale ladder and retrieval-call budget. Additional GPT-OSS retrieval-tail and question-family summaries are in Appendix Figure~\ref{fig:app_gptoss_retrieval_family}.

Table~\ref{tab:main_r1b} shows that stronger models can still exhaust the budget on some planar or flat adapters, so model scale alone is not a memory-scalability solution.
Applying the same decomposition to LoCoMo gives smaller scale-induced drops, indicating that the magnitude of the interaction burden is benchmark-dependent while the failure-source analysis remains useful (Appendix~\ref{app:locomo_external}, Table~\ref{tab:locomo_r2}).

\subsection{Finding 4: Breakdown onset gives a compact but threshold-dependent usability boundary}
\label{sec:results_breakdown}

The diagnostics above translate into different usable-scale boundaries. Under $B_0=2$ and $\alpha=0.7$, $s^*_{0.7}(A,\msys;B)$ reports the first evaluated scale where $\mathrm{Pass@}B$ falls below threshold. Table~\ref{tab:main_budget_longmemeval} reports this boundary under the primary budget and two relaxed budgets.

\begin{table}[t]
\centering
\scriptsize
\caption{\textbf{Multi-budget reliability and breakdown onset on LongMemEval.}
Each cell reports $B_0/B_1/B_2=2/3/5$ at $s_4$. Green/red marks whether the agent--memory pair stays above/falls below the 70\% reliability threshold through $s_4$.}
\label{tab:main_budget_longmemeval}
\setlength{\tabcolsep}{3pt}
\input{generated/appendix_candidates/table_app_budget_longmemeval.tex}
\end{table}

HippoRAG is unchanged across budget settings because it is single-pass, so its boundary is driven by wrong-within-budget failures rather than extra retrieval calls. LiCoMemory recovers as the budget is relaxed, but only Qwen3-32B and Qwen3-235B remain above threshold through $s_4$ across all tested budget settings; Qwen3-8B stays below threshold even at $B_2=5$. Thus, onset is not an independent score: it is a compact operational summary of the failure regime exposed by $\mathrm{Pass@}B$, $\mathrm{P90R}$, and failure-regime decomposition, memory interface, scale ladder, and retrieval-call budget.

\section{Discussion and Limitations}
The protocol supports scale-conditional claims: under a specified agent, memory interface, scale ladder, returned-unit constraint, and memory-call budget, stored evidence remains usable up to a stated reliability boundary. It does not establish a universal ranking of memory architectures, measure storage or hidden-backend efficiency, or cover stale, contradictory, privacy-sensitive, or adversarial memories. The budget is a post-hoc diagnostic over completed trajectories rather than a hard-stop deployment policy. Appendix Table~\ref{tab:reporting_card} summarizes the recommended fields for reporting scalable-memory claims.

\section{Conclusion}

We presented a scale-conditioned protocol for evaluating usable agent memory under evidence-preserving growth. The results show that storing relevant evidence is not enough: agents must access and use it through a memory interface within a stated interaction budget. By combining $\mathrm{Pass@B}$, tail retrieval-call burden, failure decomposition, and breakdown onset, the protocol makes scalable-memory claims conditional on the agent, interface, scale range, evidence setting, and budget.

\begin{ack}
This work was supported by the Jiangsu Higher Education Brand Major Construction Project (00RC029001) and the DKU Community-Based Learning Grant (00AB000026).

The authors declare no competing interests.
\end{ack}
{\small
\bibliography{references}
}

\clearpage
\appendix

\section{Appendix}
\raggedbottom

\subsection{Memory Scale Ladder}
For each query $q$ (i.e., each benchmark question/task instance), the scale level $s_k$ defines an accessible history $\mathcal{H}^{(s_k)}(q)$. This history contains the fixed evidence sessions for $q$ plus $N_{\mathrm{irr}}(s_k)$ added irrelevant sessions. The averages below are computed over queries, so ``sessions/query'' and ``tokens/query'' describe the average size of a single query's accessible history, not the total size of the benchmark.
\begin{table}[H]
  \centering
  \small
  \caption{\textbf{Memory scales} used for LongMemEval and LoCoMo. Task-relevant evidence is fixed for each query $q$; both benchmarks use the same added irrelevant-session counts $N_{\mathrm{irr}}(s_k)$ at each scale. Approximate token counts refer to the raw accessible history $\mathcal{H}^{(s_k)}(q)$ after injecting irrelevant sessions, averaged over queries.}
  \label{tab:s_longmemeval}
  \begin{tabular}{lrrrl}
  \toprule
  Scale $s_k$ & $N_{\mathrm{irr}}(s_k)$ & Avg. sessions/query & Avg. tokens/query & Added irrelevant sessions \\
  \midrule
  $s_0$  & 0   & 1.225   & $4.2$K  & evidence-only \\
  $s_1$  & 100 & 101.225 & $268$K  & +100 irrelevant sessions \\
  $s_2$  & 200 & 201.225 & $523$K  & +200 irrelevant sessions \\
  $s_3$  & 300 & 301.225 & $757$K  & +300 irrelevant sessions \\
  $s_4$  & 400 & 401.225 & $1.03$M & +400 irrelevant sessions \\
  \bottomrule
  \end{tabular}
\end{table}

\input{appendix_additions/controlled_growth}

\subsection{Evaluation Size and Scoring}
\label{app:evaluation_size}
Each main Qwen LongMemEval condition consists of 2,000 queries with a single rollout per query, resulting in 10,000 rollouts per model-memory combination across five scale levels, and a total of 60,000 rollouts across two hierarchical interfaces. The LoCoMo evaluation employs 282 queries per condition, yielding 1,410 rollouts per model-memory sweep, totaling 8,460 LoCoMo rollouts across the hierarchical interfaces. Confidence intervals are computed from 1,000 bootstrap resamples at the task level to maintain task-dependent structure.

For each trajectory, we use the LLM-as-judge for the final answer scoring.
We set $C(\tau)$ to this judgment and recompute $\mathrm{pass}_B(\tau)$, budget exhaustion, and wrong-within-budget terms from the logged retrieval-call count.
Thus, the analysis separates answer correctness from budget diagnostics, but it does not include new human relabeling or a retrieval-hit audit.

\input{appendix_additions/evaluation_details}

\subsection{LLM-as-Judge Final-Answer Scoring}
\label{app:llm_judge}

We score final answers with an LLM-as-judge protocol aligned with the public MemOS evaluation~\citep{memos}. The judge is intentionally endpoint-level: it sees the benchmark question, the gold answer, and the trajectory's generated final answer, but not the memory trajectory, retrieved evidence, or retrieval-call count. This keeps $C(\tau)$ separate from budget diagnostics such as $R(\tau)$, $p_{\mathrm{exh}}$, and $p_{\mathrm{wrong}}$.

\paragraph{Judge model and decoding.}
The judge model is \texttt{gpt-4o-mini}, called through an OpenAI-compatible chat-completions API with temperature $0$. The returned label is parsed as JSON and mapped to a binary correctness value:
\[
C(\tau)=\mathbf{1}[\texttt{label}=\texttt{"CORRECT"}].
\]
If the label is \texttt{"WRONG"}, then $C(\tau)=0$. The stored binary label is then combined with the logged retrieval-call count to compute $\mathrm{pass}_B(\tau)$.

\paragraph{System prompt.}
\begin{verbatim}
You are a careful grader. Decide whether a generated answer matches
the reference answer for a memory benchmark question.
\end{verbatim}

\noindent\begin{minipage}{\linewidth}
\paragraph{User prompt template.}
\begin{verbatim}
Grade the generated answer against the gold answer for the question.

Question:
{question}

Gold answer:
{gold_answer}

Generated answer:
{generated_answer}

Return only one JSON object:
{"label": "CORRECT"} or {"label": "WRONG"}

Mark CORRECT when the generated answer expresses the same answer as
the gold answer, even if it is longer, shorter, or phrased differently.
For date or time questions, mark CORRECT when the generated answer
refers to the same date or time period as the gold answer, even if the
format differs. Otherwise mark WRONG.
\end{verbatim}
\end{minipage}

\paragraph{Aggregation.}
The main reported trajectory label uses the parsed binary judge output for that trajectory. When repeated judge calls are available for the same final answer, we follow the MemOS metric convention and average the Boolean judge outputs at the aggregate metric level; the trajectory-level budget decomposition still uses the stored binary correctness label together with the logged retrieval-call count.

\paragraph{Compute reporting scope.}
The experiments reported here focus exclusively on inference-based evaluations without involving any training or fine-tuning of model weights. Given that some models are accessed through external model-serving services or APIs, hardware allocation and processing time from providers cannot be consistently controlled or standardized. Therefore, this study emphasizes protocol-specific resource measures, including query counts, retrieval-call budgets, scale levels, supplementary token-cost metrics, and estimated minimum local memory requirements for model serving, as detailed in Table~\ref{tab:local_compute_estimates}. Platform-dependent runtime metrics are considered separately from the main reliability analysis presented.

\begin{table}[H]
  \centering
  \small
  \caption{\textbf{Approximate local model-serving memory lower bounds.}
  Values estimate model weight memory only, using $2$ bytes/parameter for BF16/FP16, $1$ byte/parameter for 8-bit weights, and $0.5$ bytes/parameter for 4-bit weights. Practical local serving additionally requires memory for KV cache, batching/runtime overhead, retrieval indexes, and any tensor-parallel or offload buffers. For MoE models, the table reports total resident weights; active parameters per token may be smaller.}
  \label{tab:local_compute_estimates}
  \begin{tabular}{lrrrr}
    \toprule
    Backbone & Parameters & BF16/FP16 & 8-bit & 4-bit \\
    \midrule
    Qwen3-8B & 8B & $\sim$16 GB & $\sim$8 GB & $\sim$4 GB \\
    Qwen3-32B & 32B & $\sim$64 GB & $\sim$32 GB & $\sim$16 GB \\
    Qwen3-235B-A22B & 235B total / 22B active & $\sim$470 GB & $\sim$235 GB & $\sim$118 GB \\
    LLaMA-3.1-8B & 8B & $\sim$16 GB & $\sim$8 GB & $\sim$4 GB \\
    LLaMA-3.1-70B & 70B & $\sim$140 GB & $\sim$70 GB & $\sim$35 GB \\
    LLaMA-3.1-405B & 405B & $\sim$810 GB & $\sim$405 GB & $\sim$203 GB \\
    GPT-OSS-20B & 20B & $\sim$40 GB & $\sim$20 GB & $\sim$10 GB \\
    GPT-OSS-120B & 120B & $\sim$240 GB & $\sim$120 GB & $\sim$60 GB \\
    \bottomrule
  \end{tabular}
\end{table}

\input{appendix_additions/decision_examples}

\Needspace{20\baselineskip}
\subsection{Qwen Model Diagnostics: Reliability, Failure Modes, and Scalability}

\begin{table}[H]
  \centering
  \small
  \caption{\textbf{Qwen3 models used in our experiments.} We report the model name and native context length.}
  \label{tab:model-tiers}
  \begin{tabular}{ll}
    \toprule
    Model & Context \\
    \midrule
    Qwen3-8B  & 32{,}768 \\
    Qwen3-32B & 32{,}768 \\
    Qwen3-235B-A22B-Instruct-2507 & 262{,}144 \\
    \bottomrule
  \end{tabular}
\end{table}

\begin{figure*}[!htbp]
\centering
\begin{minipage}{0.49\textwidth}
\centering
\includegraphics[width=\textwidth]{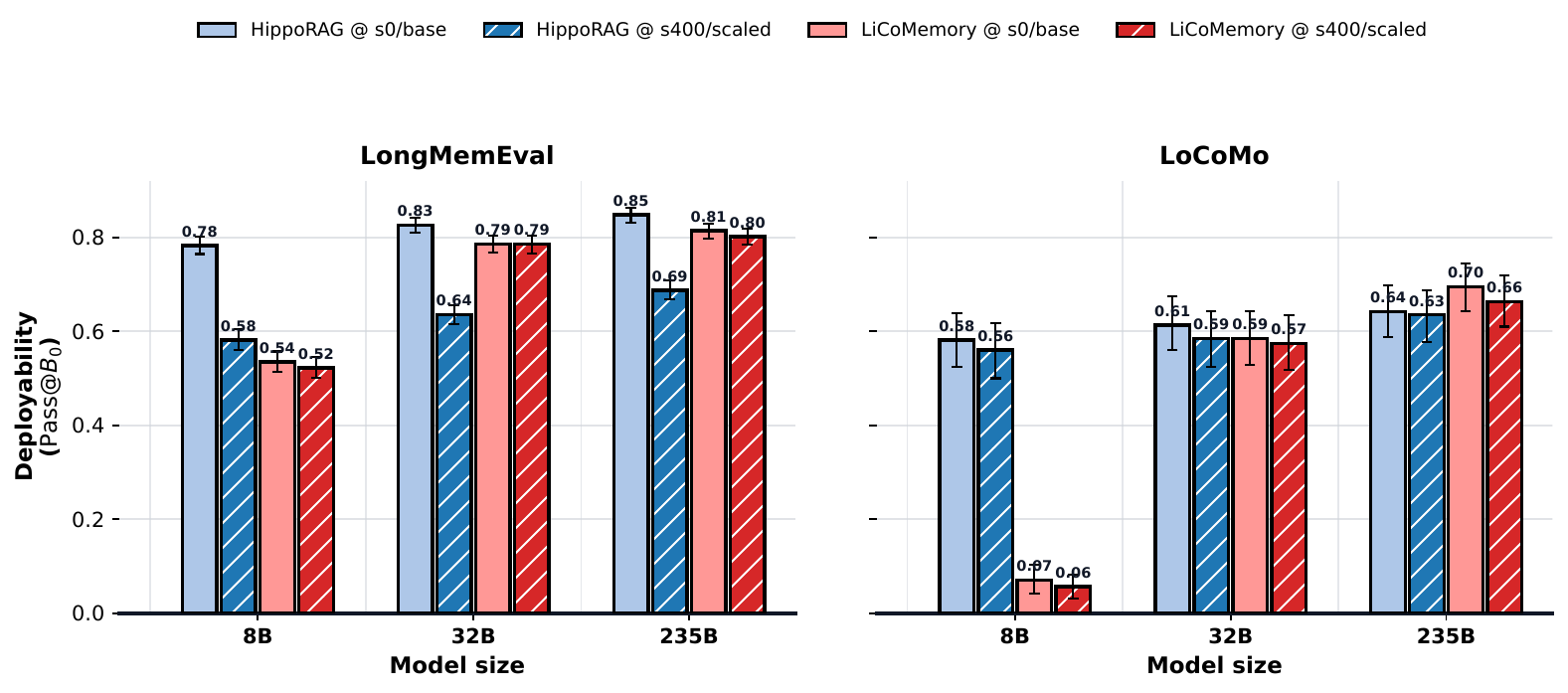}
\end{minipage}
\hfill
\begin{minipage}{0.49\textwidth}
\centering
\includegraphics[width=\textwidth]{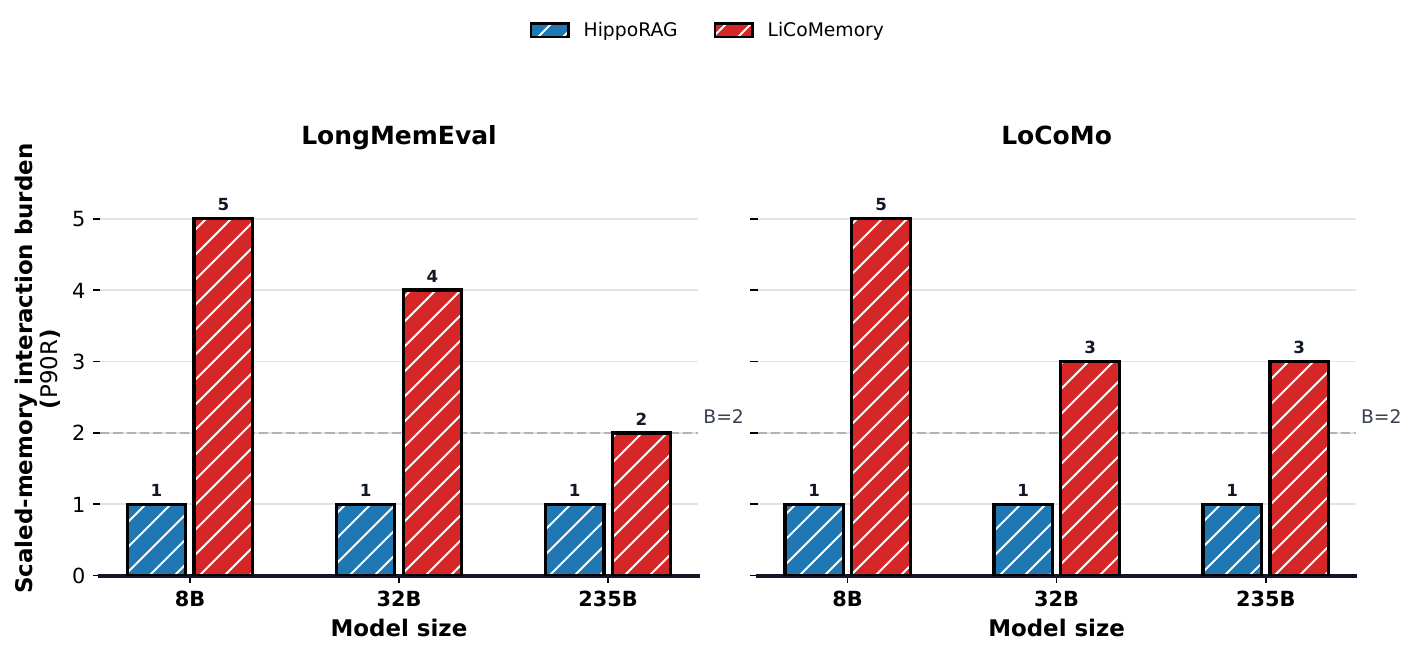}
\end{minipage}
\caption{\textbf{Qwen endpoint reliability and scaled-memory tail burden.}
Left: endpoint $\mathrm{Pass@}B_0$ (Eq.~\ref{eq:passatb}) at $s_0$ and $s_4$ under $B_0=2$.
Right: scaled-memory tail $\mathrm{P90R}(s_4)$ (Eq.~\ref{eq:burden}).}
\label{fig:app_qwen_claim_i}
\end{figure*}

Figure~\ref{fig:app_qwen_claim_i} provides supplementary visualizations of the endpoint results and tail summaries of interaction burden. The main-text Figure~\ref{fig:main_failure_decomp} visualizes the failure-regime decomposition at scaled memory for LiCoMemory. Flat and planar systems follow the same evaluation protocol, but they are not used for failure-regime analysis unless their results are explicitly reported under the same $(B_0,s)$ setting.

\paragraph{Reliability and Tail Burden Across Scale and Model Size.}
For LiCoMemory, Qwen3-8B remains below the reliability threshold at both endpoints, whereas Qwen3-32B and Qwen3-235B remain above threshold; at $s_4$, its tail retrieval count decreases from $\mathrm{P90R}=5$ to 4 and 2 across the same model tiers (Figure~\ref{fig:app_qwen_claim_i}).

\subsection{External Validity on LoCoMo}
\label{app:locomo_external}
LongMemEval provides a controlled framework by systematically introducing irrelevant sessions while maintaining fixed evidence. LoCoMo tests whether the same diagnostics remain informative in longer multi-session dialogues.

\begin{table}[H]
\centering
\small
\caption{\textbf{External validity on LoCoMo.}
Endpoint $\mathrm{Pass@}B_0$ (Eq.~\ref{eq:passatb}) is reported as mean $\pm$ 95\% CI (bootstrapped over tasks) at the shared-scale endpoints $s_0$ and $s_4$ under $B_0=2$. The scaled endpoint uses the same irrelevant-session count convention as Table~\ref{tab:s_longmemeval}, so results are comparable under matched irrelevant-session loads.}
\label{tab:locomo_r2}
\input{generated/locomo/table_r2_locomo.tex}
\end{table}

On LoCoMo, additional irrelevant sessions produce smaller drops than on LongMemEval for HippoRAG, while LiCoMemory remains strongly tied to the Qwen tier (Table~\ref{tab:locomo_r2}). Qwen3-8B stays low across both endpoints ($7.1\% \rightarrow 5.7\%$), whereas Qwen3-32B and Qwen3-235B retain higher reliability with modest endpoint declines ($58.5\% \rightarrow 57.4\%$ and $69.5\% \rightarrow 66.3\%$).

Thus, LoCoMo does not imply a fixed universal drop size. Instead, it supports the evaluation-level conclusion that the same reporting protocol remains interpretable when accessible irrelevant memory changes, with the magnitude depending on the benchmark, memory interface, and Qwen tier.

\input{appendix_additions/locomo_diagnosis}

\FloatBarrier

\subsection{Supplementary Results: Additional Model and Interface Diagnostics}\label{app:llama}

\begin{table}[H]
      \centering
      \caption{\textbf{LLaMA-3.1 models evaluated in this study.} We report the model name and native context length.}
      \label{tab:model-tiers-llama}
      \begin{tabular}{lll}
        \toprule
        Model                     & Architecture & Native context \\
        \midrule
        LLaMA-3.1-8B-Instruct     & Dense        & 128{,}000 \\
        LLaMA-3.1-70B-Instruct    & Dense        & 128{,}000 \\
        LLaMA-3.1-405B-Instruct   & Dense        & 128{,}000 \\
        \bottomrule
      \end{tabular}
\end{table}

Table~\ref{tab:model-tiers-llama} summarizes the LLaMA-3.1 model variants included in our evaluation.
This section extends the main analysis with additional model-family and interface diagnostics: LLaMA-3.1, GPT-OSS under LiCoMemory, OpenClaw, and additional flat/planar/tree adapters.

\begin{figure}[H]
  \centering
  \includegraphics[width=0.86\textwidth]{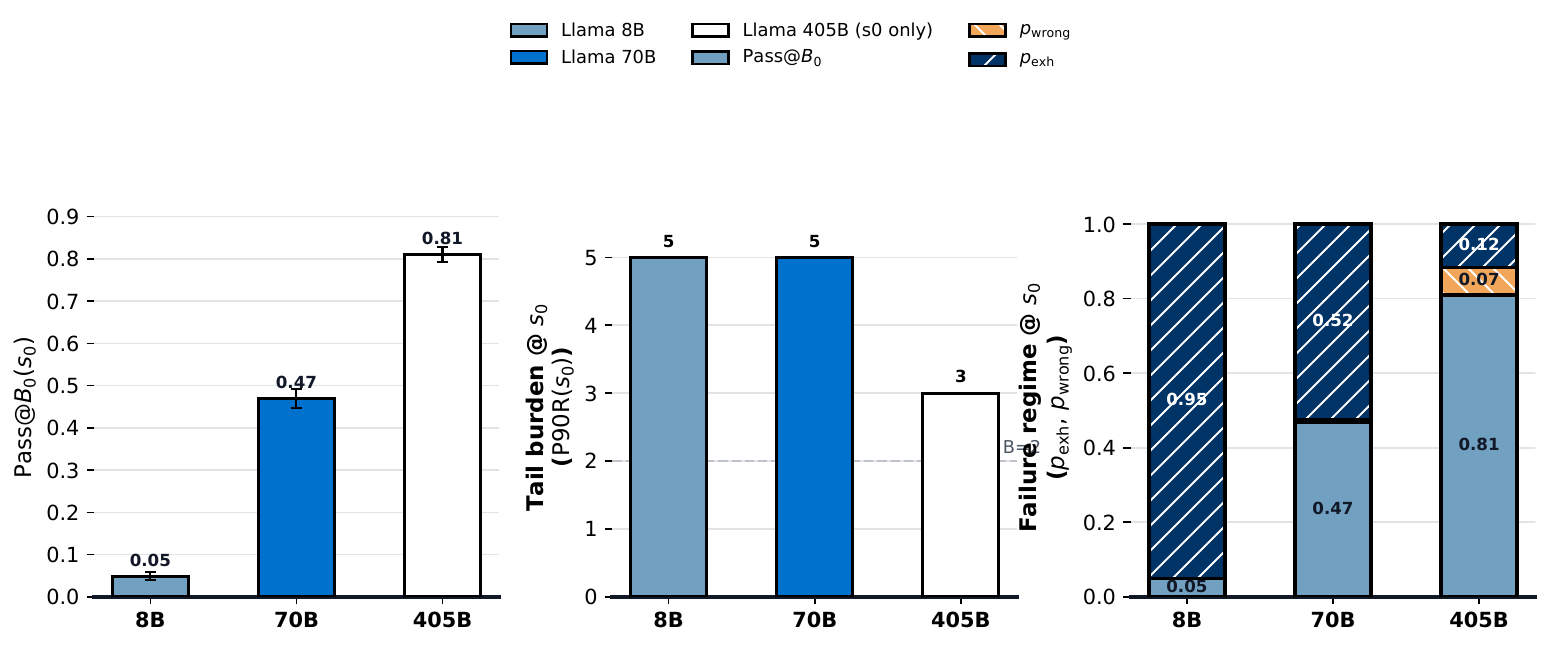}
  \caption{\textbf{LLaMA-3.1 $s_0$ baseline performance.} $\mathrm{Pass@}B_0$, interaction burden, and failure summaries for 8B, 70B, and 405B at the evidence-only setting. Hatched bars denote observed $s_0$ performance for 405B.}
  \label{fig:app_llama_405b}
\end{figure}

Figure~\ref{fig:app_llama_405b} reports baseline ($s_0$) results for all three LLaMA-3.1 models (8B, 70B, 405B). At this endpoint, larger variants have better observed $\mathrm{Pass@}B_0$ and interaction-burden profiles. The hatched bars for 405B indicate observed values, highlighting the effect of increased backbone size at the smallest scale setting.

For a direct quantitative comparison, Table~\ref{tab:app_llama_summary} presents endpoint results for the 8B and 70B models at both baseline ($s_0$) and the largest fully observed shared scale ($s_3$). These results can be directly compared to the Qwen series, illustrating both architecture- and scale-dependent trends in memory retrieval performance.

\begin{table}[H]
  \centering
  \small
  \caption{\textbf{LLaMA-3.1 endpoints on LongMemEval.} Direct comparison on the fully observed shared endpoints $s_0$ (baseline) and $s_3$ for model sizes 8B and 70B.
  Green highlights the stronger 70B rows; red marks the 8B rows with severe exhaustion at $s_0$.}
  \label{tab:app_llama_summary}
  \resizebox{0.92\columnwidth}{!}{\input{generated/llama/table_app_llama_summary.tex}}
\end{table}

\begin{figure*}[!htbp]
\centering
\begin{minipage}{0.49\textwidth}
\centering
\includegraphics[width=\textwidth]{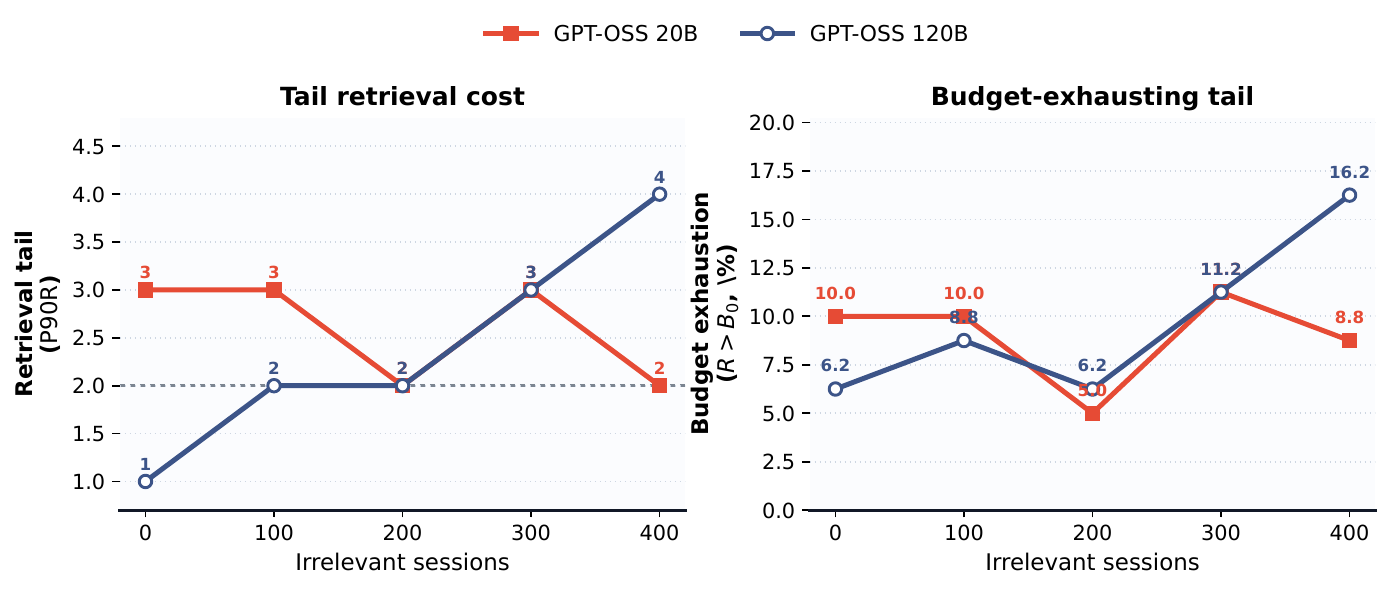}
\end{minipage}
\hfill
\begin{minipage}{0.49\textwidth}
\centering
\includegraphics[width=\textwidth]{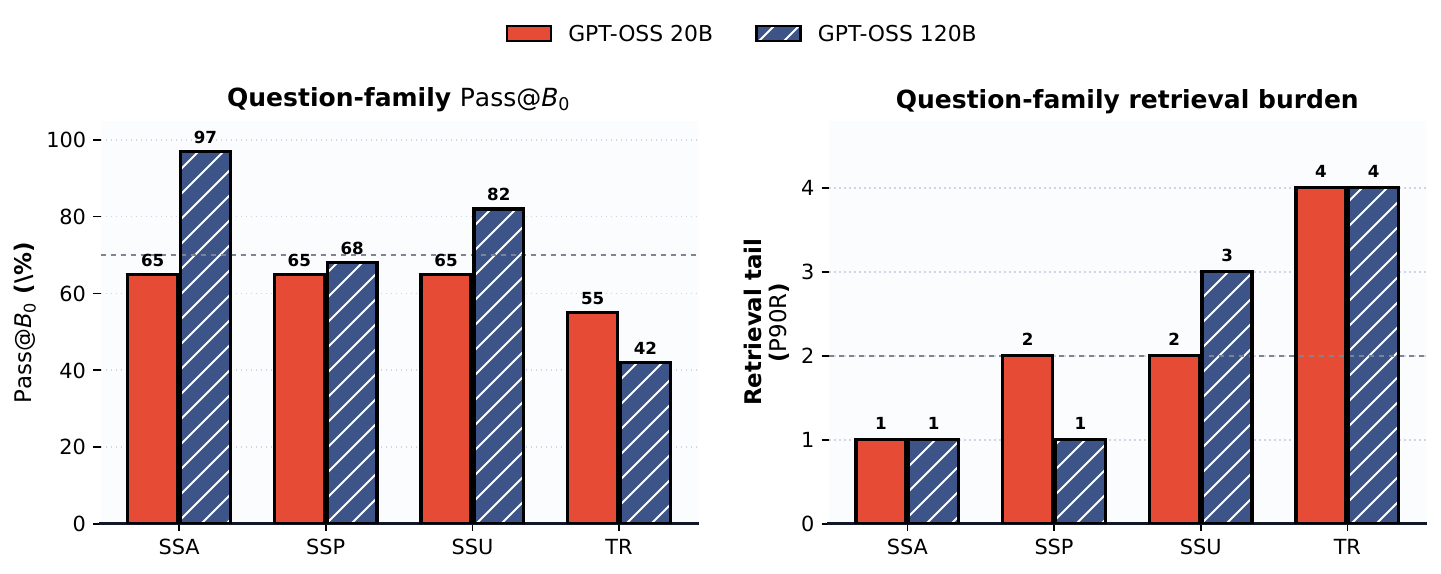}
\end{minipage}
\caption{\textbf{GPT-OSS LiCoMemory retrieval and question-family diagnostics.}
Left: retrieval-tail behavior across scale, including $\mathrm{P90R}$ and the fraction of budget-induced rollouts.
Right: question-family $\mathrm{Pass@}B_0$ and retrieval-tail summaries.}
\label{fig:app_gptoss_retrieval_family}
\end{figure*}

We also report GPT-OSS~\citep{gptoss} LiCoMemory retrieval-tail and question-family diagnostics in Figure~\ref{fig:app_gptoss_retrieval_family}, complementing the scale and failure-regime decomposition in Figure~\ref{fig:gptoss_licomemory_main}.

We additionally report OpenClaw diagnostics on LongMemEval. These figures mirror the same diagnostic chain used in the main text: scale-level reliability, output decomposition, interaction burden, and supplementary cost accounting.

\begin{figure*}[!htbp]
\centering
\begin{minipage}{0.49\textwidth}
\centering
\includegraphics[width=\textwidth]{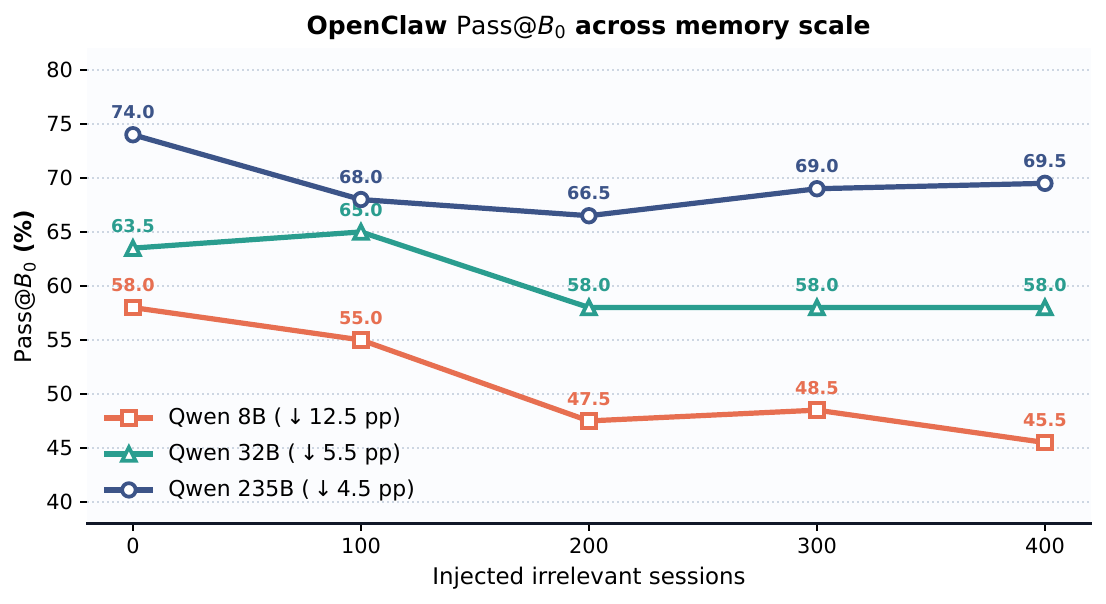}
\end{minipage}
\hfill
\begin{minipage}{0.49\textwidth}
\centering
\includegraphics[width=\textwidth]{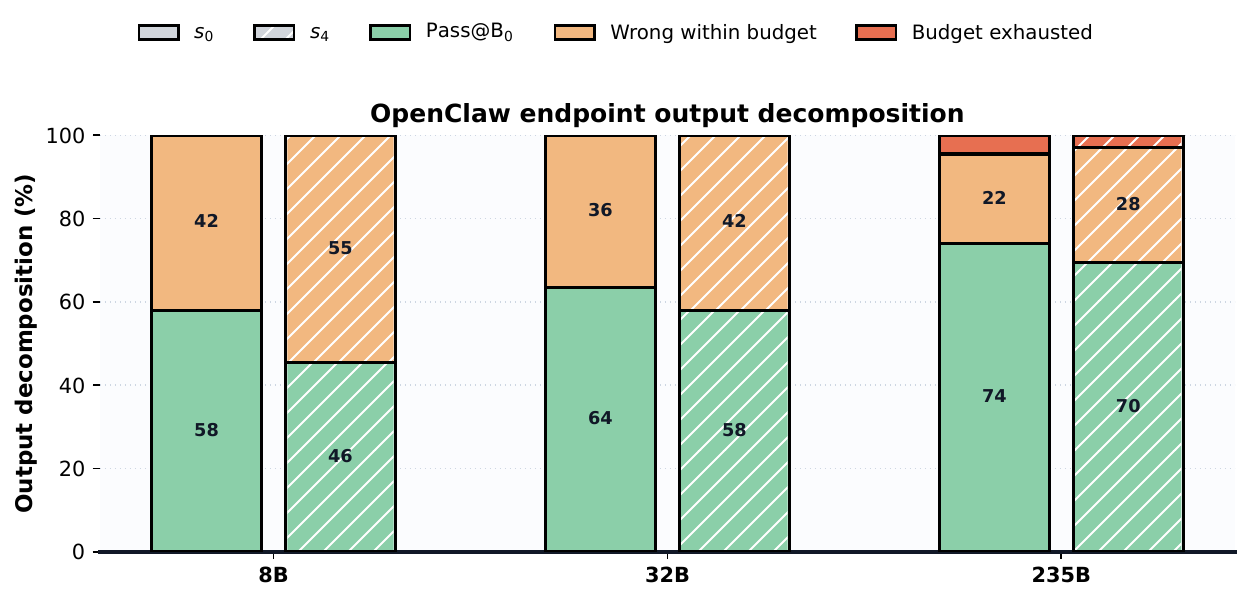}
\end{minipage}
\caption{\textbf{OpenClaw reliability and output decomposition on LongMemEval.}
Left: $\mathrm{Pass@}B_0$ across the shared memory-scale ladder.
Right: endpoint output decomposition at $s_0$ and $s_4$ into $\mathrm{Pass@}B_0$, within-budget wrong answers, and budget-induced trajectories.}
\label{fig:app_openclaw_scale_output}
\end{figure*}

\begin{figure*}[!htbp]
\centering
\begin{minipage}{0.49\textwidth}
\centering
\includegraphics[width=\textwidth]{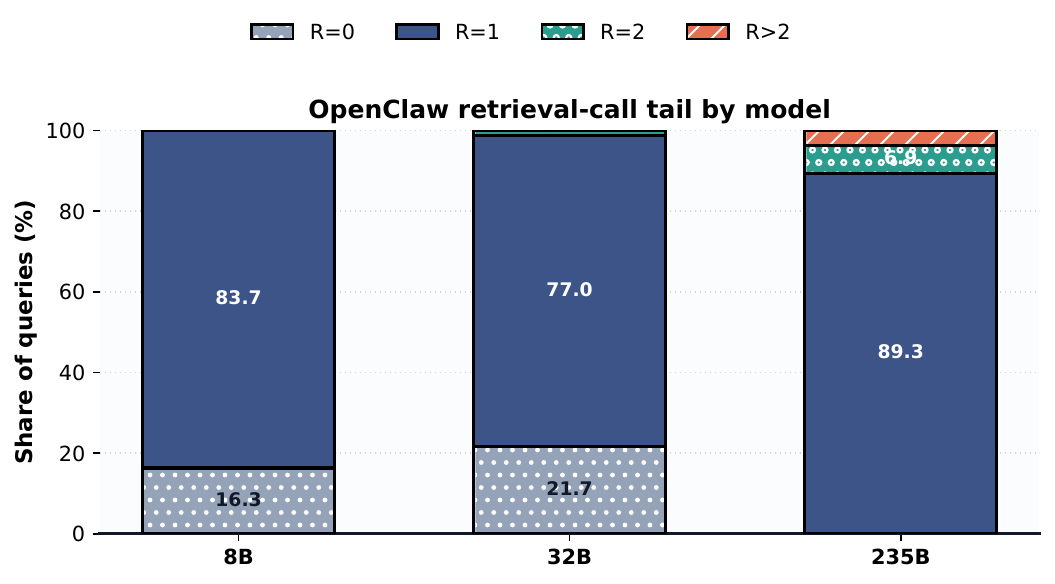}
\end{minipage}
\hfill
\begin{minipage}{0.49\textwidth}
\centering
\includegraphics[width=\textwidth]{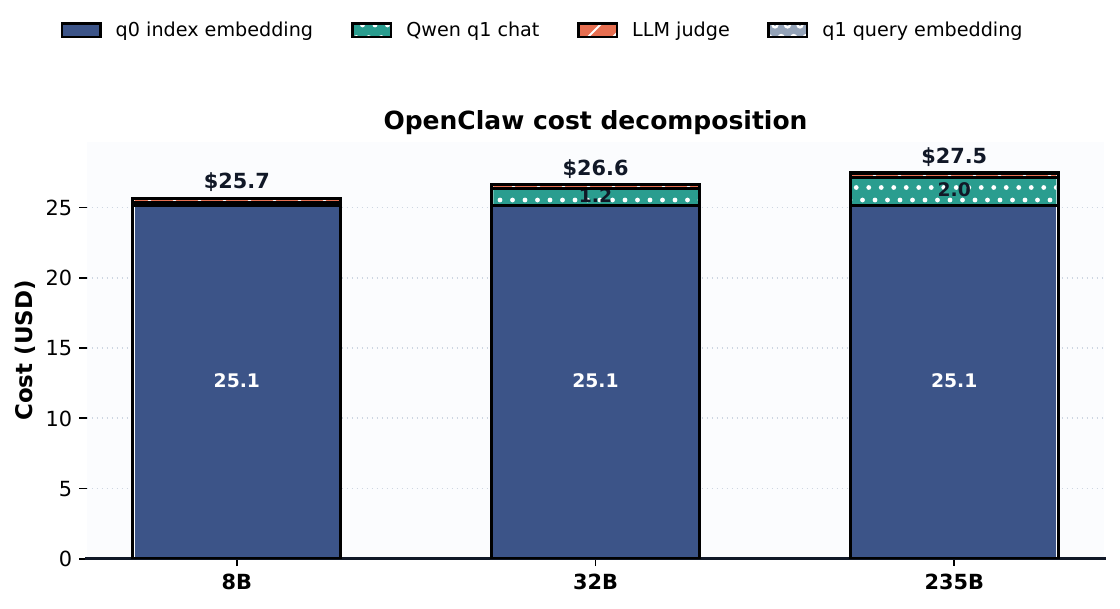}
\end{minipage}
\caption{\textbf{OpenClaw interaction burden and cost decomposition.}
Left: retrieval-call distribution by model, including the share of queries exceeding $B_0=2$.
Right: cost decomposition separated into indexing, query-time chat, query embedding, and judging cost.}
\label{fig:app_openclaw_tail_cost}
\end{figure*}

\input{appendix_additions/additional_systems}

\input{appendix_additions/controller_sensitivity}

\FloatBarrier
\subsection{Additional Diagnostics}
\label{app:diagnostics}

\noindent These diagnostics provide additional support for the main claims and make the post-processing logic explicit. Table~\ref{tab:reporting_card} summarizes the information to report when evaluating memory scalability.

\begin{table}[H]
\centering
\scriptsize
\caption{\textbf{Information to report when evaluating memory scalability.}}
\label{tab:reporting_card}
\setlength{\tabcolsep}{4pt}
\begin{tabularx}{0.92\columnwidth}{lX}
\toprule
Field & Example reported in this paper \\
\midrule
Agent and interface & Qwen3 agents with flat, planar, and hierarchical memory interfaces \\
Scale range & $0$--$400$ added irrelevant sessions \\
Evidence condition & evidence-preserving: annotated evidence sessions fixed \\
Retrieval budget & $B_0=2$ agent-issued memory calls; sensitivity at $B\in\{3,5\}$ \\
Reliability & $\mathrm{Pass@B}$ under the stated scale and budget \\
Interaction burden & $\mathrm{P90R}$, the tail of agent-issued memory calls \\
Failure regime & budget-induced vs. wrong-within-budget \\
Breakdown onset & first evaluated scale below threshold $\alpha$ \\
\bottomrule
\end{tabularx}
\end{table}

\paragraph{Budget IDs and multi-budget appendix diagnostics.}
We report $B_0=2$ as the primary budget. Larger budgets are used only as supplementary diagnostics to quantify how much reliability can be recovered when the retrieval-call budget is relaxed under the same memory-scale ladder and returned-item constraints. Table~\ref{tab:budgets} defines the named budget IDs used throughout the paper; the corresponding LongMemEval scaled-endpoint and breakdown summaries for the two hierarchical interfaces are reported in Table~\ref{tab:main_budget_longmemeval}.

\begin{table}[H]
  \centering
  \small
  \caption{\textbf{Retrieval-call budget IDs used in this paper.} $B_0$ is the primary budget in the main results; larger budgets are used for appendix diagnostics.}
  \label{tab:budgets}
  \begin{tabular}{lr}
    \toprule
    Budget ID & Value \\
    \midrule
    $B_0$ (main)    & 2 \\
    $B_1$ (elastic) & 3 \\
    $B_2$ (stress)  & 5 \\
    \bottomrule
  \end{tabular}
\end{table}

\begin{figure}[H]
\centering
\begin{minipage}{0.49\textwidth}
\centering
\includegraphics[width=\textwidth]{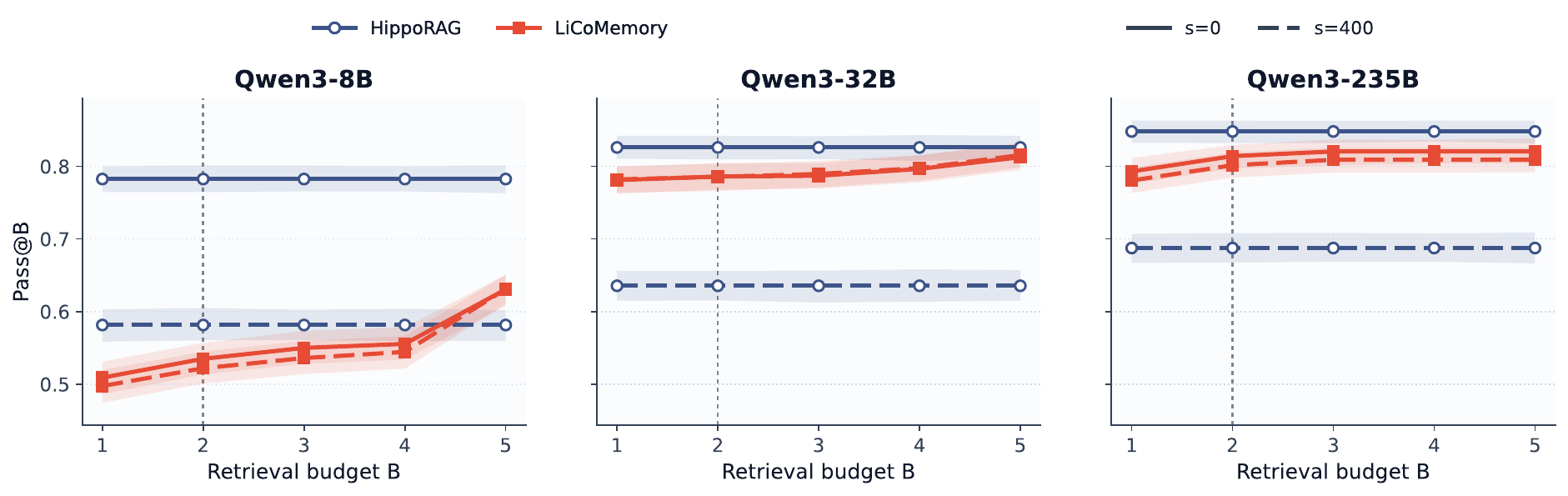}
\end{minipage}
\hfill
\begin{minipage}{0.49\textwidth}
\centering
\includegraphics[width=\textwidth]{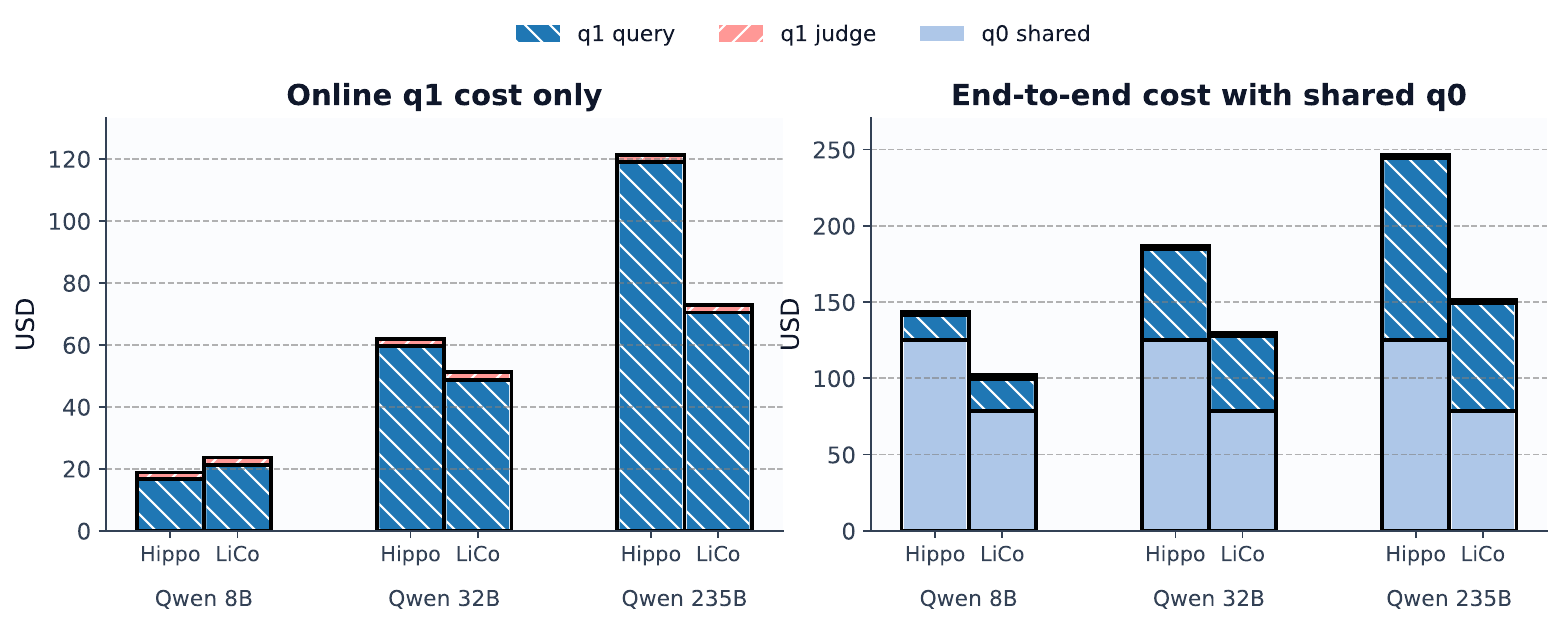}
\end{minipage}
\caption{\textbf{Additional diagnostics on representative LongMemEval settings.}
Left: reliability curves $\mathrm{Pass@}B$ under $B \in \{1,\dots,5\}$.
HippoRAG is nearly budget-insensitive in these curves because the evaluated adapter exposes a single agent-visible retrieval call, whereas LiCoMemory recovers as the budget is relaxed, especially for smaller models.
Right: q0/q1 cost accounting.
This panel separates shared preprocessing cost from online query-time cost and clarifies that the paper's retrieval-budget analysis targets the online agent-facing burden rather than denying offline cost differences across systems.}
\label{fig:app_budget_and_cost}
\end{figure}

\paragraph{Supplementary token-cost accounting.}
The core protocol uses a retrieval-call budget rather than a token budget. We therefore report token expenditure only as a supplementary systems diagnostic in this appendix. In particular, the right panel of Figure~\ref{fig:app_budget_and_cost} separates online query-time tokens ($q1$) from amortized preprocessing-plus-query tokens ($q0{+}q1$), where $q0$ denotes preprocessing/indexing cost and $q1$ denotes online query-time cost. This lets cost placement in the pipeline be inspected without conflating token accounting with the reliability metric $\mathrm{Pass@}B$.

\begin{figure}[H]
\centering
\includegraphics[width=0.76\textwidth]{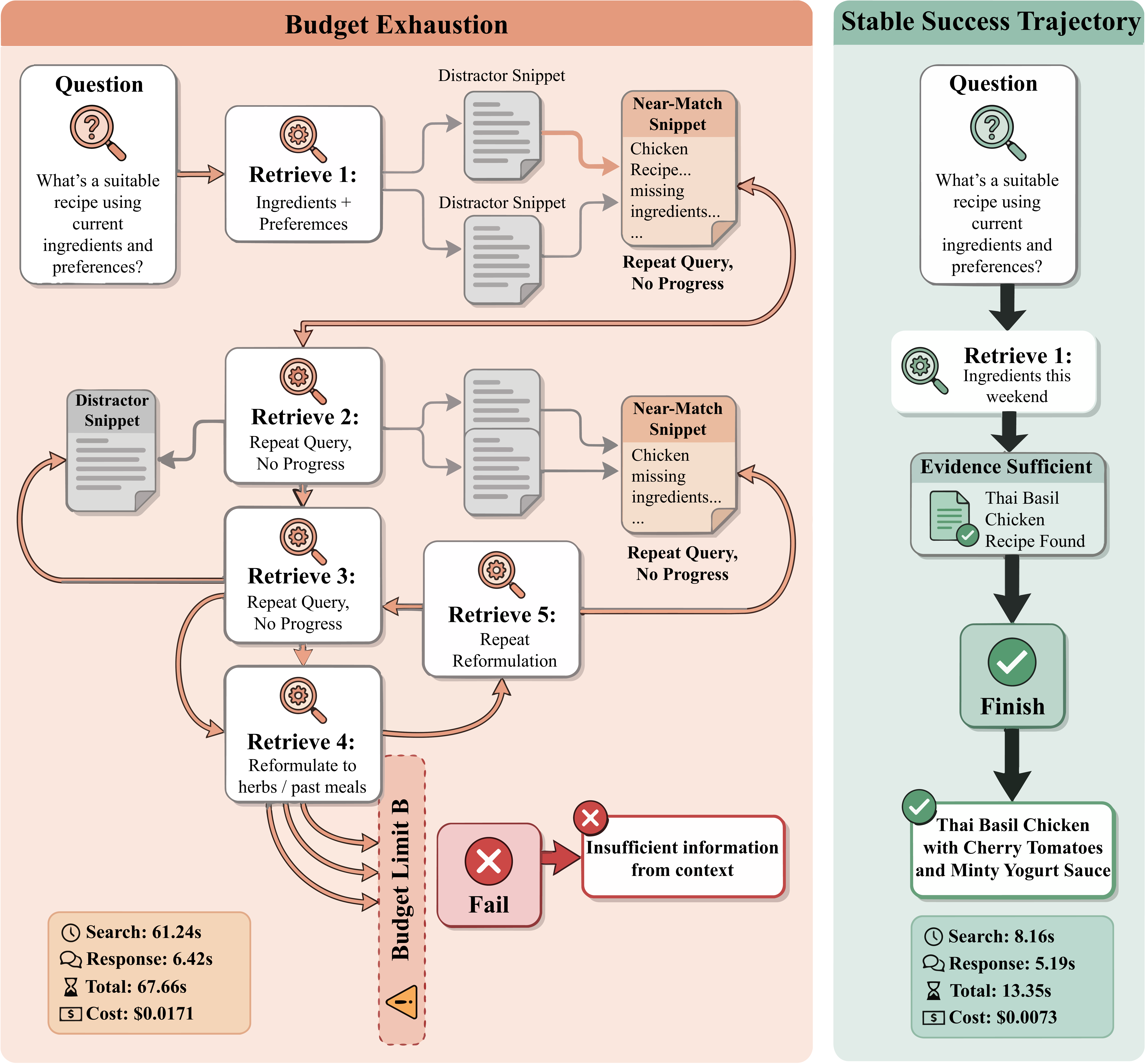}
\caption{\textbf{Matched qualitative trajectories on the same LongMemEval item under LiCoMemory at $s=400$.}
The weaker model violates the retrieval-call budget after repeated retrieve--reformulate steps, whereas the stronger model reaches the answer after one effective retrieval on the same item.}
\label{fig:qualitative_pair}
\end{figure}

\paragraph{Qualitative evidence source.}
Figure~\ref{fig:qualitative_pair} provides a representative matched LongMemEval pair under LiCoMemory at $s=400$.

\FloatBarrier
\subsection{Detailed Memory Taxonomy}
\label{app:taxonomy}

\begin{table}[H]
\centering
\caption{\textbf{Detailed memory taxonomy used by the unified evaluation protocol.}
Flat, planar, and hierarchical labels are assigned by the agent-visible retrieval interface used in our evaluation protocol, and systems are evaluated under the same scale ladder and retrieval-call budget; when reported, we also match the number of model-visible returned evidence units per retrieval step across systems.
System references: OpenClaw~\citep{openclaw_memory_overview}, MemOS variants~\citep{memos}, Mem0~\citep{mem0}, HippoRAG~\citep{hipporag}, and LiCoMemory~\citep{licomemory}.}
\label{tab:memory-taxonomy-full}
{\small
\setlength{\tabcolsep}{4pt}
\begin{tabularx}{\textwidth}{
>{\raggedright\arraybackslash}p{0.4cm}
>{\raggedright\arraybackslash}p{2.0cm}
>{\raggedright\arraybackslash}p{2.0cm}
>{\raggedright\arraybackslash}X
>{\raggedright\arraybackslash}X
>{\raggedright\arraybackslash}p{2.0cm}
}
\toprule
ID & Method & Subtype & Store unit (granularity) & Retrieve & Update Policy \\
\midrule
F1 & OpenClaw & Flat document memory &
retrieved document/text block &
single-pass top-$k$ document-memory retrieval &
not modeled in the reported endpoint runs \\
F2 & MemOS-text & Flat text memory &
chunk / memory cube text &
hybrid retrieval over flat textual memories &
merge / memory maintenance \\
P1 & Mem0 & Iterative memory search &
session or chunk memory item &
agent-controlled repeated semantic memory queries before final answer &
incremental add / update \\
P2 & MemOS-Tree & Planarized tree memory &
memory-tree node exposed through a single tree lookup layer &
tree-structured retrieval over textual memories &
tree maintenance / merge \\
H1 & HippoRAG & Hierarchical graph &
multi-granularity graph memory (passage / fact / entity) &
single-pass hybrid retrieval with graph propagation &
graph augmentation / refresh \\
H2 & LiCoMemory & Cognigraph (hierarchical graph) &
entity nodes + relation edges with chunk/session provenance &
entity-led graph retrieval + temporal-semantic reranking &
incremental node/edge merge \\
\bottomrule
\end{tabularx}
}
\end{table}

\clearpage
\subsection{Flat and Planar Parity Conditions}
\label{app:fp_parity}

The protocol is defined for flat, planar, and hierarchical memory interfaces. To avoid conflating memory scale with implementation-specific access differences, any flat or planar run reported under this protocol must satisfy the same parity constraints as the hierarchical runs. Specifically, for each query $q$ and scale $s$, the accessible history $\mathcal{H}^{(s)}$ must contain the same task-relevant evidence and the same injected irrelevant sessions; the retrieval-call budget must be $B_0=2$ for primary results; the scale levels must be $s_0$--$s_4$ with $N_{\mathrm{irr}}\in\{0,100,200,300,400\}$; and each retrieval action must return the same number of model-visible evidence units. For dense or single-pass systems this is the returned top-$k$ list; for graph or multi-stage systems it is the final set returned after internal traversal or reranking. Systems may differ in chunking, indexing, linking, traversal, reranking, and update policy, but not in the underlying evidence, irrelevant-session set, or per-query retrieval-call budget.

Under these constraints, the reported flat systems (OpenClaw and MemOS-text), planar systems (Mem0 and MemOS-Tree), and hierarchical systems (HippoRAG and LiCoMemory) are evaluated as different memory interfaces over the same agent-facing memory-scale condition. MemOS-Tree is treated as planarized in our adapter because the agent observes a single tree-retrieval layer rather than a multi-level abstraction policy. Therefore, comparisons across F/P/H systems should be interpreted as interface comparisons, not as changes in task difficulty or memory-store content.

\end{document}

%% file: generated/longmemeval/table_r1_longmemeval.tex
{\renewcommand{\arraystretch}{0.88}%
\begin{tabular}{llcccc}
\toprule
\multirow{2}{*}{Agent $A$} & \multirow{2}{*}{Memory $\msys$} & \multicolumn{2}{c}{Pass@$B_0$ (\%) $\uparrow$} & P90R & Onset $s^*_{0.7}$ \\
\cmidrule(lr){3-4}
 & & $s_0$ & $s_4$ & @$s_4$ $\downarrow$ & (irr.\ sessions) $\uparrow$ \\
\midrule
\multirow{2}{*}{Qwen3-8B} & HippoRAG & \textbf{78.2 $\pm$ 1.9} & \textbf{58.2 $\pm$ 2.3} ({\textcolor[HTML]{990000}{$\downarrow$20.0}}) & \textbf{1} & \textbf{100} \\
 & LiCoMemory & 53.5 $\pm$ 2.2 & 52.2 $\pm$ 2.3 ({$\downarrow$1.3}) & \textcolor[HTML]{990000}{5} & 0 \\
\midrule
\multirow{2}{*}{Qwen3-32B} & HippoRAG & \textbf{82.6 $\pm$ 1.7} & 63.5 $\pm$ 2.1 ({\textcolor[HTML]{990000}{$\downarrow$19.1}}) & \textbf{1} & 100 \\
 & LiCoMemory & 78.6 $\pm$ 1.9 & \textbf{78.5 $\pm$ 1.9} ({$\downarrow$0.1}) & \textcolor[HTML]{990000}{4} & \textbf{>400} \\
\midrule
\multirow{2}{*}{Qwen3-235B} & HippoRAG & \textbf{84.8 $\pm$ 1.6} & 68.8 $\pm$ 2.0 ({\textcolor[HTML]{990000}{$\downarrow$16.0}}) & \textbf{1} & 300 \\
 & LiCoMemory & 81.4 $\pm$ 1.7 & \textbf{80.1 $\pm$ 1.7} ({$\downarrow$1.3}) & 2 & \textbf{>400} \\
\bottomrule
\end{tabular}}

%% file: generated/longmemeval/table_r1b_longmemeval_failure.tex
{\setlength{\tabcolsep}{4pt}%
  \renewcommand{\arraystretch}{0.82}%
  \setlength{\fboxsep}{0.8pt}%
  \begin{tabular}{@{}llcc@{\hspace{10pt}}cc@{}}
  \toprule
  \multicolumn{2}{c}{} & \multicolumn{2}{c}{\textbf{Budget-noncompliant} ($p_{\mathrm{exh}}$, \%)} & \multicolumn{2}{r}{\textbf{Wrong-within-budget}
  ($p_{\mathrm{wrong}}$, \%)} \\
  \cmidrule(lr){3-4}\cmidrule(lr){5-6}
  Agent model $A$ & Memory system $\msys$ & $s_0$ & $s_4$ & $s_0$ & $s_4$ \\
  \midrule
  \multirow{9}{*}{Qwen3-8B}
   & \multicolumn{5}{l}{\textit{Flat}} \\
   & \quad OpenClaw & 0.0 & 0.0 & 42.0 & \colorbox[HTML]{F4CCCC}{\textbf{54.5}} \\
   & \quad MemOS-text & 2.6 & 0.4 & 45.8 & \colorbox[HTML]{F4CCCC}{\textbf{97.2}} \\
   & \multicolumn{5}{l}{\textit{Planar}} \\
   & \quad Mem0 & 4.2 & 1.4 & 52.2 & \colorbox[HTML]{F4CCCC}{\textbf{85.6}} \\
   & \quad MemOS-Tree & 23.0 & 30.2 & 36.0 & \colorbox[HTML]{F4CCCC}{\textbf{68.2}} \\
   & \multicolumn{5}{l}{\textit{Hierarchical}} \\
   & \quad HippoRAG & 0.0 & 0.0 & 21.8 & \colorbox[HTML]{F4CCCC}{\textbf{41.8}} \\
   & \quad LiCoMemory & 37.2 & \colorbox[HTML]{F4CCCC}{\textbf{38.6}} & 9.3 & 9.2 \\
  \midrule
  \multirow{9}{*}{Qwen3-32B}
   & \multicolumn{5}{l}{\textit{Flat}} \\
   & \quad OpenClaw & 0.0 & 0.0 & 36.5 & \colorbox[HTML]{F4CCCC}{\textbf{42.0}} \\
   & \quad MemOS-text & 41.2 & \colorbox[HTML]{F4CCCC}{\textbf{69.0}} & 17.2 & 28.2 \\
   & \multicolumn{5}{l}{\textit{Planar}} \\
   & \quad Mem0 & 46.0 & \colorbox[HTML]{F4CCCC}{\textbf{83.6}} & 16.4 & 9.4 \\
   & \quad MemOS-Tree & 3.4 & 4.4 & 54.6 & \colorbox[HTML]{F4CCCC}{\textbf{93.2}} \\
   & \multicolumn{5}{l}{\textit{Hierarchical}} \\
   & \quad HippoRAG & 0.0 & 0.0 & 17.4 & \colorbox[HTML]{F4CCCC}{\textbf{36.5}} \\
   & \quad LiCoMemory & 15.4 & \colorbox[HTML]{F4CCCC}{\textbf{15.7}} & 6.0 & 5.8 \\
  \midrule
  \multirow{9}{*}{Qwen3-235B}
   & \multicolumn{5}{l}{\textit{Flat}} \\
   & \quad OpenClaw & 4.5 & 3.0 & 21.5 & \colorbox[HTML]{F4CCCC}{\textbf{27.5}} \\
   & \quad MemOS-text & 51.8 & \colorbox[HTML]{F4CCCC}{\textbf{79.6}} & 9.2 & 6.2 \\
   & \multicolumn{5}{l}{\textit{Planar}} \\
   & \quad Mem0 & 46.0 & \colorbox[HTML]{F4CCCC}{\textbf{79.2}} & 30.8 & 4.0 \\
   & \quad MemOS-Tree & 12.6 & \colorbox[HTML]{F4CCCC}{\textbf{57.4}} & 19.0 & 16.0 \\
   & \multicolumn{5}{l}{\textit{Hierarchical}} \\
   & \quad HippoRAG & 0.0 & 0.0 & 15.2 & \colorbox[HTML]{F4CCCC}{\textbf{31.2}} \\
   & \quad LiCoMemory & 1.9 & 2.0 & 16.8 & \colorbox[HTML]{F4CCCC}{\textbf{17.9}} \\
  \bottomrule
  \end{tabular}
  }

%% file: generated/appendix_candidates/table_app_budget_longmemeval.tex
{\renewcommand{\arraystretch}{0.88}%
\begin{tabularx}{\columnwidth}{ll>{\centering\arraybackslash}X>{\centering\arraybackslash}X}
\toprule
Agent $A$ & Memory $\msys$ & Pass@$B$$(s_4)$ (\%) $\uparrow$ & Onset $s^*_{0.7}$ (irr.\ sessions) $\uparrow$ \\
\midrule
\multirow{2}{*}{Qwen3-8B} & HippoRAG & 58.1 / 58.1 / 58.1 & 100 / 100 / 100 \\
 & LiCoMemory & \cellcolor[HTML]{F4CCCC}52.2 / 53.6 / 63.1 & \cellcolor[HTML]{F4CCCC}0 / 0 / 0 \\
\midrule
\multirow{2}{*}{Qwen3-32B} & HippoRAG & 63.5 / 63.5 / 63.5 & 100 / 100 / 100 \\
 & LiCoMemory & \cellcolor[HTML]{E2F0D9}78.5 / 79.0 / 81.5 & \cellcolor[HTML]{E2F0D9}\textbf{>400 / >400 / >400} \\
\midrule
\multirow{2}{*}{Qwen3-235B} & HippoRAG & 68.8 / 68.8 / 68.8 & 300 / 300 / 300 \\
 & LiCoMemory & \cellcolor[HTML]{E2F0D9}80.2 / 80.9 / 80.9 & \cellcolor[HTML]{E2F0D9}\textbf{>400 / >400 / >400} \\
\bottomrule
\end{tabularx}}

%% file: appendix_additions/controlled_growth.tex
\paragraph{Why preserve evidence during growth?}
Irrelevant sessions are complete conversations, not random tokens. They are not part of the annotated evidence for the current query, but can still be topically similar, compete for retrieval, and occupy returned context. Holding the evidence and gold answer fixed lets us test the effect of irrelevant-session growth without changing what is needed to answer the query. In deployment, history size can change together with evidence validity, redundancy, and user preferences. Table~\ref{tab:app_memory_interventions} separates these variables. For example, if a user's preference changes while the history grows, a previously correct answer may become wrong even when retrieval succeeds. The controlled ladder avoids mixing that change with the effect of irrelevant-session growth.

\begin{table}[H]
  \centering
  \small
  \caption{\textbf{Memory interventions change different variables.} The scale ladder varies irrelevant-session count while preserving the query's evidence and gold answer.}
  \label{tab:app_memory_interventions}
  \begin{tabular}{lll}
    \toprule
    Intervention & Variable changed & Evaluation task \\
    \midrule
    Irrelevant sessions & Accessible history size & Find fixed evidence \\
    Duplicate evidence & Evidence redundancy & Find repeated support \\
    Stale evidence & Evidence validity & Identify current facts \\
    Conflicting evidence & Evidence consistency & Resolve disagreement \\
    Updated preferences & Target state & Answer for the new state \\
    \bottomrule
  \end{tabular}
\end{table}

%% file: appendix_additions/evaluation_details.tex
\paragraph{Supplementary system coverage.}
Table~\ref{tab:app_qwen_coverage} gives the sample sizes for the Qwen analyses. Each row covers Qwen3-8B, Qwen3-32B, and Qwen3-235B at all five scales. MemOS-text, Mem0, and MemOS-Tree use the same 500-query set, giving 22,500 trajectories across 45 conditions. LiCoMemory's curves use 2,000 queries under the same scale construction, so the four-system plot is not a paired comparison on 500 queries. The categories group the evaluated implementations by interface, while each result describes a specific agent--implementation pair.

\begin{table}[H]
  \centering
  \small
  \caption{\textbf{Qwen trajectory coverage.} A condition fixes the benchmark, system, agent, and scale. Counts exclude the separate step-cap reruns.}
  \label{tab:app_qwen_coverage}
  \begin{tabular}{llrr}
    \toprule
    Benchmark & Systems & Per condition & Total \\
    \midrule
    LongMemEval & HippoRAG, LiCoMemory & 2,000 & 60,000 \\
    LongMemEval & MemOS-text, Mem0, MemOS-Tree & 500 & 22,500 \\
    LongMemEval & OpenClaw & 200 & 3,000 \\
    LoCoMo & HippoRAG, LiCoMemory & 282 & 8,460 \\
    \bottomrule
  \end{tabular}
\end{table}

\paragraph{Decoding and metric reproduction.}
The Qwen agents and final-answer judge use temperature $0$. Task-level bootstrap intervals quantify variation across queries, not repeated-generation variability for the same query. We focus on system-selection reversals, distinct failure partitions, and changes of many percentage points rather than interpreting sub-point differences. The public audit tables retain each trajectory's task identifier, benchmark, system, model, scale, correctness label, and agent-visible call count. The script \texttt{reproduce\_metrics.py} in \texttt{artifacts/reviewer3\_audit/scripts/} reconstructs the coverage and four diagnostics from these records. The call-distribution and failure-decomposition plots both use the $s_4$ trajectories.

%% file: appendix_additions/decision_examples.tex
\clearpage
\subsection{Decision-Changing Examples}
\label{app:decision_examples}

The diagnostics answer three questions that a single accuracy score leaves open: which system to choose under a call budget, which failure type accounts for a low score, and over what scale range the reliability target is met. The examples below use the recorded correctness labels and call counts. ``Success'' means that the completed trajectory's final answer is correct, regardless of whether it meets the call budget.

\paragraph{Example 1: system selection.}
Table~\ref{tab:app_selection_reversal} compares HippoRAG and LiCoMemory with Qwen3-8B on the same LongMemEval queries and $s_4$ histories. Final-answer success favors LiCoMemory, $63.10\%$ versus $58.15\%$. Requiring a correct answer within two calls instead favors HippoRAG, $58.15\%$ versus $52.20\%$. At a five-call budget, LiCoMemory is again preferred. Each system is scored from its own fixed set of completed trajectories. Only the reporting criterion changes, not the agent's behavior.

\begin{table}[H]
  \centering
  \small
  \caption{\textbf{System selection under different call requirements.} LongMemEval, Qwen3-8B, $s_4$, with 2,000 trajectories per system. All entries are percentages. Bold marks the higher score in each column.}
  \label{tab:app_selection_reversal}
  \input{generated/appendix_rebuttal/selection.tex}
\end{table}

For LiCoMemory, $10.90$ percentage points of final-answer success come from trajectories exceeding two calls. An accuracy score counts these as successes, whereas $\mathrm{Pass@}2$ excludes them because they do not meet the two-call requirement. This changes which system has the higher score under that requirement.

\paragraph{Example 2: similar scores, different failure regimes.}
With Qwen3-32B at $s_4$, MemOS-text and MemOS-Tree have similarly low $\mathrm{Pass@}2$, but markedly different failure partitions (Table~\ref{tab:app_similar_scores}). MemOS-text exceeds two calls on $69.0\%$ of queries and has $\mathrm{P90R}=6$. MemOS-Tree instead ends wrong within budget on $93.2\%$ of queries and has $\mathrm{P90R}=2$. Thus, the dominant failures are long trajectories for MemOS-text and incorrect answers within two calls for MemOS-Tree. The budgeted score alone does not distinguish them.

\begin{table}[H]
  \centering
  \small
  \caption{\textbf{Similar budgeted reliability, different failure regimes.} LongMemEval, Qwen3-32B, $s_4$, with 500 trajectories per system. Rates are percentages and $\mathrm{P90R}$ is a call count. Bold marks the dominant failure term in each row.}
  \label{tab:app_similar_scores}
  \input{generated/appendix_rebuttal/similar_scores.tex}
\end{table}

\paragraph{Example 3: baseline usability and breakdown onset.}
For Qwen3-8B on LongMemEval, LiCoMemory changes from $53.50\%$ to $52.20\%$ in $\mathrm{Pass@}2$ between $s_0$ and $s_4$. Although this curve is nearly flat, it is already below the target $\alpha=0.7$ at $s_0$, giving breakdown onset $s^*_{0.7}=0$. HippoRAG starts at $78.25\%$ and first falls below the target after 100 irrelevant sessions are introduced. Breakdown onset therefore distinguishes a configuration already below target at baseline from one that meets the target initially but falls below it as history grows. The change in score alone misses this difference.

%% file: generated/appendix_rebuttal/selection.tex
\begin{tabular}{lrrrr}
\toprule
System & Success & $\mathrm{Pass@}2$ & $\mathrm{Pass@}3$ & $\mathrm{Pass@}5$ \\
\midrule
HippoRAG & 58.15 & \textbf{58.15} & \textbf{58.15} & 58.15 \\
LiCoMemory & \textbf{63.10} & 52.20 & 53.60 & \textbf{63.10} \\
\bottomrule
\end{tabular}

%% file: generated/appendix_rebuttal/similar_scores.tex
\begin{tabular}{lrrrr}
\toprule
System & $\mathrm{Pass@}2$ & $p_{\mathrm{wrong}}$ & $p_{\mathrm{exh}}$ & $\mathrm{P90R}$ \\
\midrule
MemOS-text & 2.8 & 28.2 & \textbf{69.0} & 6 \\
MemOS-Tree & 2.4 & \textbf{93.2} & 4.4 & 2 \\
\bottomrule
\end{tabular}

%% file: generated/locomo/table_r2_locomo.tex
\begin{tabular}{llccc}
\toprule
\multirow{2}{*}{Agent $A$} & \multirow{2}{*}{Memory $\msys$} & \multicolumn{2}{c}{Pass@B$_0$ (\%) $\uparrow$} & Drop \\
\cmidrule(lr){3-4}
 & & $s_0$ & $s_4$ & (pp) $\downarrow$ \\
\midrule
\multirow{2}{*}{Qwen3-8B} & HippoRAG & \textbf{58.2 $\pm$ 5.7} & \textbf{56.0 $\pm$ 6.0} & 2.1 \\
 & LiCoMemory & \cellcolor[HTML]{F4CCCC}7.1 $\pm$ 3.2 & \cellcolor[HTML]{F4CCCC}5.7 $\pm$ 2.5 & \textbf{1.4} \\
\midrule
\multirow{2}{*}{Qwen3-32B} & HippoRAG & \textbf{61.3 $\pm$ 6.0} & \textbf{58.5 $\pm$ 6.0} & 2.8 \\
 & LiCoMemory & 58.5 $\pm$ 5.7 & 57.4 $\pm$ 6.0 & \textbf{1.1} \\
\midrule
\multirow{2}{*}{Qwen3-235B} & HippoRAG & 64.2 $\pm$ 5.7 & 63.5 $\pm$ 5.7 & \textbf{0.7} \\
 & LiCoMemory & \cellcolor[HTML]{E2F0D9}\textbf{69.5 $\pm$ 5.3} & \cellcolor[HTML]{E2F0D9}\textbf{66.3 $\pm$ 5.7} & 3.2 \\
\bottomrule
\end{tabular}

%% file: appendix_additions/locomo_diagnosis.tex
\Needspace{30\baselineskip}
\paragraph{Baseline usability and interaction burden.}
Table~\ref{tab:app_locomo_diagnosis} reports final-answer success, budgeted reliability, and call counts together for LiCoMemory. With Qwen3-8B, final-answer success is $41.49\%$ at both $s_0$ and $s_4$. At $s_0$, however, only $7.09\%$ of trajectories succeed within two calls, and $91.84\%$ exceed two calls. The $34.40$-percentage-point gap between success and $\mathrm{Pass@}2$ consists of correct final answers from over-budget trajectories. Adding 400 irrelevant sessions reduces $\mathrm{Pass@}2$ by $1.42$ points. The low budgeted score is therefore already present before irrelevant-session growth, rather than arising from it.

\begin{table}[H]
  \centering
  \small
  \setlength{\tabcolsep}{5pt}
  \caption{\textbf{LiCoMemory on LoCoMo: correctness and call counts.} Each row contains 282 trajectories. Rates are percentages, $B=2$, and $\mathrm{P90R}$ is a call count. Success denotes final-answer correctness without the call criterion.}
  \label{tab:app_locomo_diagnosis}
  \input{generated/appendix_rebuttal/locomo_diagnosis.tex}
\end{table}

The larger Qwen agents have substantially higher baseline $\mathrm{Pass@}2$ and fewer over-budget trajectories with the same LiCoMemory implementation. Reporting these metrics together separates each configuration's baseline behavior from its response to history growth. It also shows whether failures under the two-call requirement are mainly wrong answers within budget or trajectories that exceed it. Neither category alone identifies which internal component caused the failure.

%% file: generated/appendix_rebuttal/locomo_diagnosis.tex
\begin{tabular}{llrrrrr}
\toprule
Agent & Scale & Success & $\mathrm{Pass@}2$ & $p_{\mathrm{wrong}}$ & $p_{\mathrm{exh}}$ & $\mathrm{P90R}$ \\
\midrule
Qwen3-8B & $s_0$ & 41.49 & 7.09 & 1.06 & 91.84 & 5 \\
 & $s_4$ & 41.49 & 5.67 & 0.35 & 93.97 & 5 \\
Qwen3-32B & $s_0$ & 68.44 & 58.51 & 14.18 & 27.30 & 4 \\
 & $s_4$ & 67.02 & 57.45 & 15.96 & 26.60 & 3 \\
Qwen3-235B & $s_0$ & 74.47 & 69.50 & 17.73 & 12.77 & 3 \\
 & $s_4$ & 69.50 & 66.31 & 20.92 & 12.77 & 3 \\
\bottomrule
\end{tabular}

%% file: generated/llama/table_app_llama_summary.tex
\begin{tabular}{lcccc}
\toprule
Model & Pass@B$_0$@$s_0$ (\%) & Pass@B$_0$@$s_3$ (\%) & Cost@$s_0$ (med/p90) & Exhaustion@$s_0$ (\%) \\
\midrule
Llama 8B  & \cellcolor[HTML]{F4CCCC}\textbf{4.9} & \cellcolor[HTML]{F4CCCC}\textbf{4.7} & 5 / 5 & \cellcolor[HTML]{F4CCCC}\textbf{95.1} \\
Llama 70B & \cellcolor[HTML]{E2F0D9}\textbf{46.9} & \cellcolor[HTML]{E2F0D9}\textbf{46.9} & 4 / 5 & 52.5 \\
\bottomrule
\end{tabular}

%% file: appendix_additions/additional_systems.tex
\FloatBarrier
\subsection{Scale-Conditioned Diagnostics across Additional Memory Systems}
\label{app:additional_systems}

We evaluate MemOS-text, Mem0, and MemOS-Tree using the same diagnostics: reliability across scale, failure decomposition, call burden, and recovery under a larger call budget. The analysis contains 22,500 LongMemEval trajectories, with 500 queries per system, Qwen tier, and scale. Each system's configuration stays fixed as irrelevant sessions are added.

\begin{figure}[!htbp]
  \centering
  \includegraphics[width=0.98\textwidth]{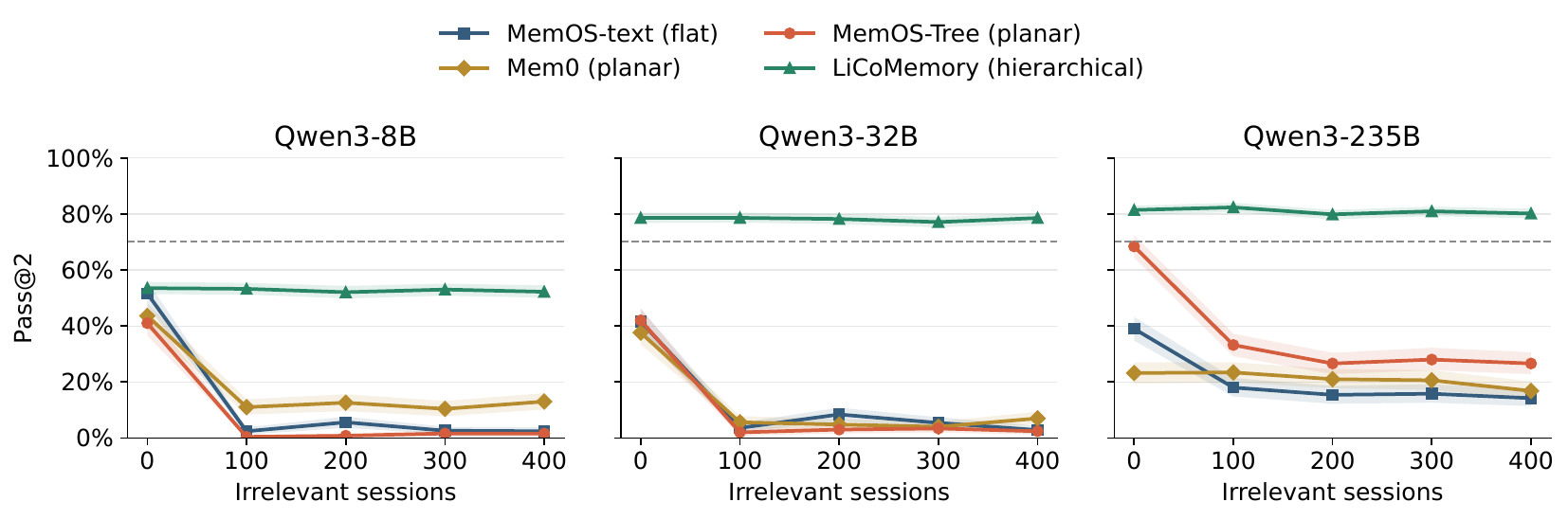}
  \caption{\textbf{Reliability across scale for four memory implementations.} All panels use LongMemEval and $B_0=2$. Each point contains 500 trajectories for MemOS-text, Mem0, and MemOS-Tree, and 2,000 for LiCoMemory. Shading denotes task-bootstrap 95\% intervals. The dashed line marks $\alpha=0.7$.}
  \label{fig:app_memory_system_interface_trends}
\end{figure}

\paragraph{Reliability across scale.}
Figure~\ref{fig:app_memory_system_interface_trends} compares four implementations spanning the paper's three categories: flat MemOS-text, planar Mem0 and MemOS-Tree, and hierarchical LiCoMemory. MemOS-Tree has a tree-structured backend, but exposes a single lookup layer to the agent (Table~\ref{tab:memory-taxonomy-full}). For MemOS-text, Mem0, and MemOS-Tree, $\mathrm{Pass@}2$ declines as irrelevant sessions accumulate. LiCoMemory retains higher budgeted reliability with Qwen3-32B and Qwen3-235B. The curves compare these specific agent--implementation pairs, not all systems in an interface category.

\begin{figure}[!htbp]
  \centering
  \includegraphics[width=0.94\textwidth]{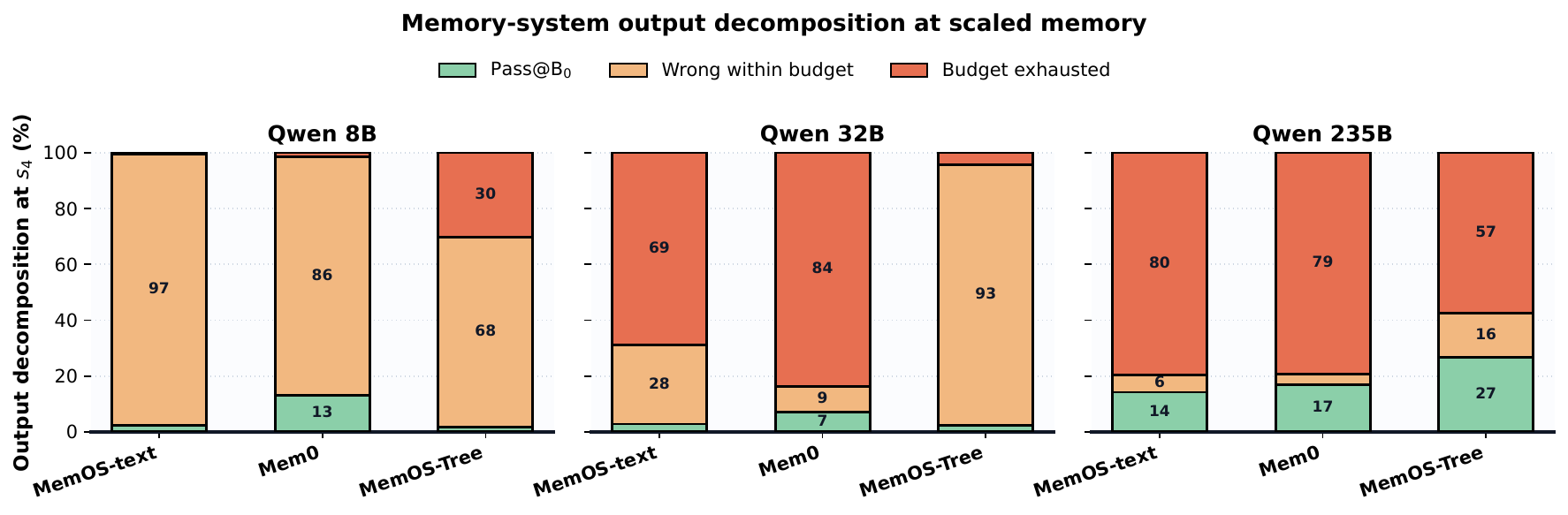}
  \caption{\textbf{Failure partition at scaled memory.} For each system and Qwen tier, 500 trajectories at $s_4$ are partitioned into correct-within-budget, wrong-within-budget, and over-budget outcomes under $B_0=2$.}
  \label{fig:app_memory_system_output_decomp}
\end{figure}

\paragraph{Failure decomposition and call burden.}
Figure~\ref{fig:app_memory_system_output_decomp} separates wrong-within-budget answers from trajectories exceeding two calls at $s_4$. The distinction depends on both the memory implementation and the agent. For example, Mem0 with Qwen3-8B ends wrong within budget on $85.6\%$ of queries, whereas Mem0 with Qwen3-32B exceeds the budget on $83.6\%$. The corresponding call distributions in Figure~\ref{fig:app_memory_system_retrieval_tail} make this behavioral difference visible. The similar-score example in Appendix~\ref{app:decision_examples} illustrates the same diagnostic value across two implementations.

\begin{figure}[!htbp]
  \centering
  \includegraphics[width=0.98\textwidth]{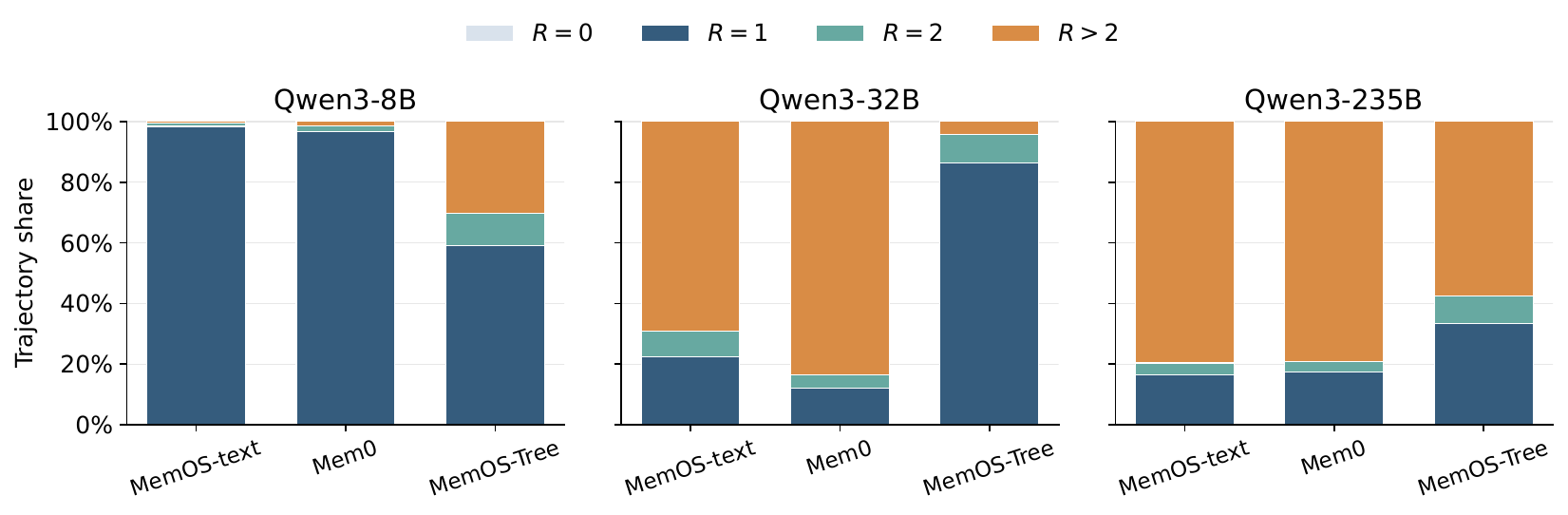}
  \caption{\textbf{Call distributions for the same scaled-memory conditions.} Each bar contains the 500 $s_4$ trajectories used in Figure~\ref{fig:app_memory_system_output_decomp}. The $R>2$ share equals $p_{\mathrm{exh}}$ at $B_0=2$.}
  \label{fig:app_memory_system_retrieval_tail}
\end{figure}

\begin{figure}[H]
  \centering
  \includegraphics[width=0.86\textwidth]{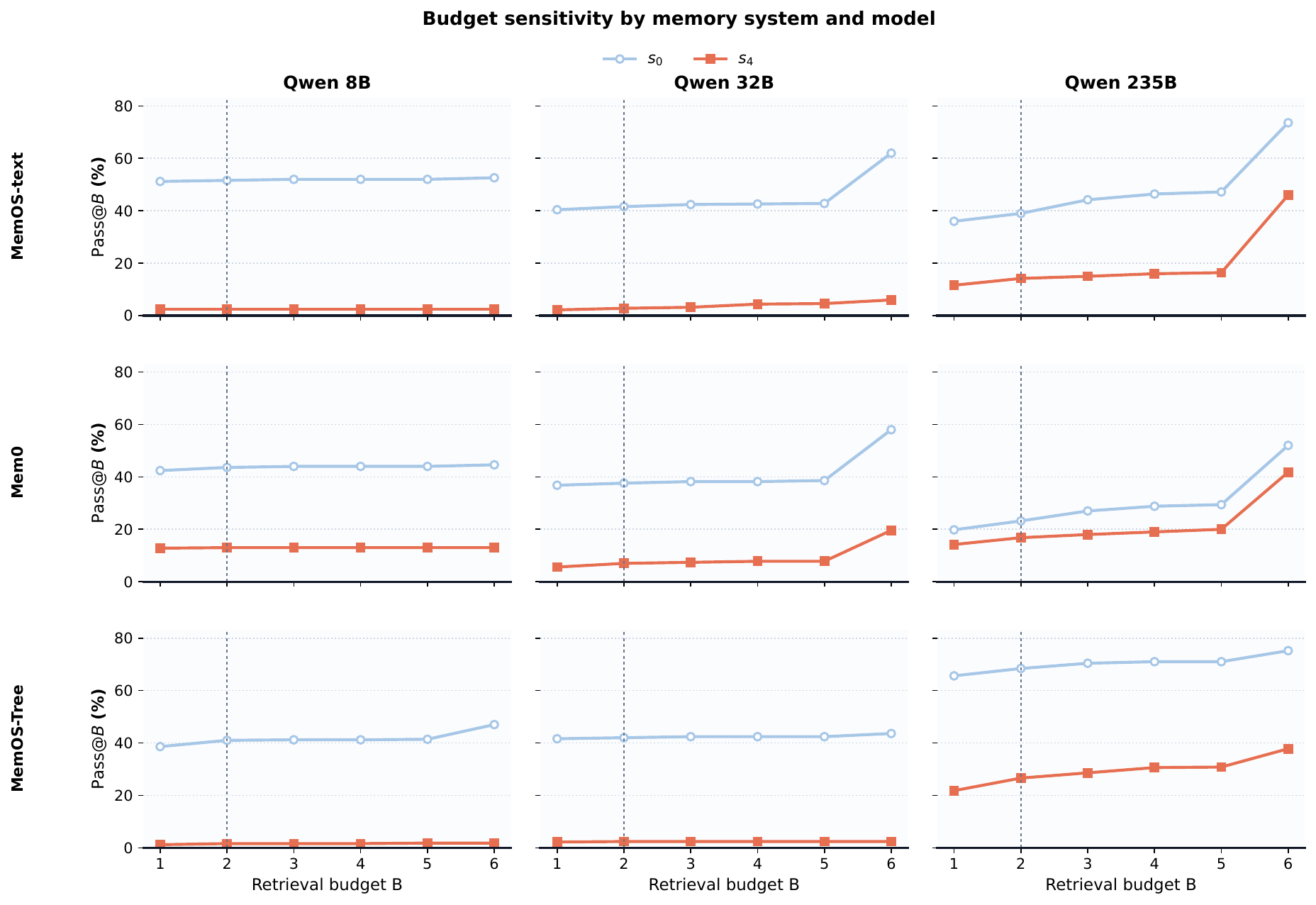}
  \caption{\textbf{Budget sensitivity across systems and agents.} Each panel compares $s_0$ and $s_4$ as the post-hoc call criterion increases, with the completed trajectories fixed. Recovery beyond $B=2$ counts correct final answers from longer trajectories.}
  \label{fig:app_memory_system_budget_sensitivity}
\end{figure}

\FloatBarrier
\paragraph{Does a larger budget recover correct answers?}
Figure~\ref{fig:app_memory_system_budget_sensitivity} scores the same trajectories at budgets from one to six calls. Table~\ref{tab:app_tree_recovery} summarizes MemOS-Tree. At $s_4$, increasing the reporting budget from two to six calls raises $\mathrm{Pass@}B$ by $0.2$ percentage points for Qwen3-8B and none for Qwen3-32B, compared with $11.2$ points for Qwen3-235B. The larger agent has both higher two-call reliability and more correct answers among its longer trajectories. For the smaller agents, few additional trajectories are correct even with the larger allowance. The budget curves show how many correct final answers fall outside the two-call limit but satisfy a larger call budget.

\begin{table}[H]
  \centering
  \small
  \caption{\textbf{MemOS-Tree reliability and budget recovery.} All entries are percentages, with 500 trajectories per condition.}
  \label{tab:app_tree_recovery}
  \input{generated/appendix_rebuttal/tree_budget_recovery.tex}
\end{table}

\FloatBarrier

%% file: generated/appendix_rebuttal/tree_budget_recovery.tex
\begin{tabular}{lrrr}
\toprule
Agent & $\mathrm{Pass@}2(s_0)$ & $\mathrm{Pass@}2(s_4)$ & $\mathrm{Pass@}6(s_4)$ \\
\midrule
Qwen3-8B & 41.0 & 1.6 & 1.8 \\
Qwen3-32B & 42.0 & 2.4 & 2.4 \\
Qwen3-235B & 68.4 & 26.6 & 37.8 \\
\bottomrule
\end{tabular}

%% file: appendix_additions/controller_sensitivity.tex
\Needspace{30\baselineskip}
\subsection{ReAct Step-Cap Sensitivity and Post-Hoc Budget Evaluation}
\label{app:controller_sensitivity}

The evaluation budget $B$ is applied to completed trajectories. It is separate from the step limit used to generate them. We examine this distinction with LiCoMemory on LongMemEval. From the original runs with a lower ReAct step limit, we select every trajectory with exactly two retrieval calls. We rerun these queries with a six-step ReAct limit, allowing the agent to finish on its own and using no external controller. This gives 4,485, 1,798, and 1,517 reruns for Qwen3-8B, Qwen3-32B, and Qwen3-235B, respectively, totaling 7,800. Each rerun replaces its original record, while all other records remain unchanged. Table~\ref{tab:app_controller_sensitivity} compares the full sets before and after this replacement: 10,000 trajectories per agent across five scales, not just the rerun subset.

\begin{table}[H]
  \centering
  \small
  \setlength{\tabcolsep}{4pt}
  \caption{\textbf{ReAct step-cap sensitivity under a fixed evaluation budget.} Each entry shows before $\to$ after replacing the original $R=2$ records with six-step reruns. Rates are percentages pooled over five LongMemEval scales. Both sets are scored at $B=2$.}
  \label{tab:app_controller_sensitivity}
  \input{generated/appendix_rebuttal/controller_comparison.tex}
\end{table}

For Qwen3-8B, final-answer success increases by $1.71$ percentage points, while wrong-within-budget answers decrease from $38.48\%$ to $9.55\%$ and over-budget trajectories increase from $0\%$ to $37.65\%$. The final-answer scores are close, but the failure partition changes substantially. Reporting both reveals a behavioral change that the accuracy score alone largely conceals.

By contrast, the multi-budget analysis holds the generation setup fixed and applies $B=2,3,5$ to the same completed trajectories. It measures how many outcomes meet each call requirement without changing the agent's policy. Giving a different remaining-search prompt for each budget would also change the policy, so the comparison would no longer isolate the effect of the reporting criterion. The experiment above varies the ReAct step limit, not such a prompt. This limit counts ReAct action turns, whereas $B$ counts agent-issued memory calls.

%% file: generated/appendix_rebuttal/controller_comparison.tex
\begin{tabular}{lrrrr}
\toprule
Agent & Success & $\mathrm{Pass@}2$ & $p_{\mathrm{wrong}}$ & $p_{\mathrm{exh}}$ \\
\midrule
Qwen3-8B & $61.52 \to 63.23$ & $61.52 \to 52.80$ & $38.48 \to 9.55$ & $0.00 \to 37.65$ \\
Qwen3-32B & $79.96 \to 81.00$ & $79.96 \to 78.21$ & $20.04 \to 6.18$ & $0.00 \to 15.61$ \\
Qwen3-235B & $80.97 \to 81.80$ & $80.97 \to 80.94$ & $19.03 \to 16.85$ & $0.00 \to 2.21$ \\
\bottomrule
\end{tabular}